\documentclass[twoside,11pt]{article}
\usepackage{jair, theapa, rawfonts}

\usepackage[implicit=false]{hyperref}

\jairheading{76}{2023}{1305-1342}{10/2022}{04/2023}
\ShortHeadings{QNLP in Practice: Compositional Models of Meaning on a Quantum Computer}
{Lorenz, Pearson, Meichanetzidis, Kartsaklis \& Coecke}
\firstpageno{1305}

\usepackage[dvipsnames]{xcolor}

\usepackage{latexsym}
\usepackage{amsmath}
\usepackage{tikz}

\usepackage{ stmaryrd }
\usepackage{float}

\usepackage{enumitem}
\usepackage{braket}
\usepackage{subcaption}
\usepackage{amssymb}

\newcommand{\MC}{{\textit{MC}}}
\newcommand{\RP}{{\textit{RP}}}
\newcommand{\discocat}{\textsc{DisCoCat}}
\newcommand{\ws}{word-sequence}
\newcommand{\bow}{bag-of-words}
\newcommand{\ansaetze}{ans{\"a}tze}
\newcommand{\Ansaetze}{Ans{\"a}tze}
\newcommand{\MCModel}{$(1, 3, 1)$}
\newcommand{\RPModel}{$(0, 1, 2)$}

\newcommand{\ed}[1]{\textcolor{black}{#1}} 
\newcommand{\rl}[1]{\textcolor{black}{#1}}

\title{QNLP in Practice: Running Compositional Models of Meaning on a Quantum Computer}

\author{\name Robin Lorenz \email robin.lorenz@quantinuum.com \\
       \name Anna Pearson \email anna.pearson@quantinuum.com \\
       \name Konstantinos Meichanetzidis \email k.mei@quantinuum.com \\
       \name Dimitri Kartsaklis \email dimitri.kartsaklis@quantinuum.com \\
       \name Bob Coecke \email bob.coecke@quantinuum.com \\
       \addr Quantinuum LLC\\ 
       17 Beaumont Street, Oxford, OX1 2NA, UK}

\pgfdeclarelayer{edgelayer}
\pgfdeclarelayer{nodelayer}
\pgfsetlayers{edgelayer,nodelayer,main}

\tikzstyle{none}=[inner sep=0pt, thick]
\tikzstyle{plain}=[inner sep=0pt, thick]
\tikzstyle{every picture}=[baseline=(current bounding box).east, scale=0.3, node distance=5mm, text height=5pt, text depth=0pt]

\tikzstyle{black_dot}=[fill=black, draw=black, shape=circle, inner sep=0pt, minimum size=0.1cm, text height=2pt, text depth=0pt]
\tikzstyle{white_dot}=[fill=none, draw=black, shape=circle, inner sep=0pt, minimum size=0.2cm]

\tikzstyle{dashed_line}=[dashed, gray, line width=0.25mm]

\newcommand{\ctikzfig}[1]{%
\begin{center}\rm
   
\InputIfFileExists{#1.tikz}{}{\input{.//tikz_figures//#1.tikz}}

\end{center}}

\newcommand{\newcite}[1]{\citeauthor{#1} \citeyear{#1}}

\begin{document}
\maketitle
\begin{abstract}
Quantum Natural Language Processing (QNLP) deals with the design and implementation of NLP models intended to be run on quantum hardware. In this paper, we present results on the first  NLP experiments conducted on Noisy Intermediate-Scale Quantum (NISQ) computers for datasets of size greater than 100 sentences. Exploiting the formal similarity of the compositional model of meaning by \newcite{CoeckeSadrClark2010} with quantum theory, we create representations for sentences that have a natural mapping to quantum circuits. We use these representations to implement and successfully train NLP models that solve simple sentence classification tasks on quantum hardware. We conduct quantum simulations that compare the syntax-sensitive model of Coecke et al. with two baselines that use less or no syntax; specifically, we implement the quantum analogues of a ``bag-of-words'' model, where syntax is not taken into account at all, and of a word-sequence model, where only word order is respected. 
We demonstrate that all models converge smoothly both in simulations and when run on quantum hardware, and that the results are the expected ones based on the nature of the tasks and the datasets used. Another important goal of this paper is to describe in a way accessible to AI and NLP researchers the main principles, process and challenges of experiments on quantum hardware. Our aim in doing this is to take the first small steps in this unexplored research territory and pave the way for practical Quantum Natural Language Processing.
\end{abstract}

\section{Introduction}
\label{sec:intro}

With the potential to provide computational speedups over the current standard, quantum computing has rapidly evolved to become one of the most popular cutting-edge areas in computer science, with promising results and applications spanning a wide range of topics such as cryptography \shortcite{Pirandola_2020}, chemistry \shortcite{Cao2019}, and biomedicine \shortcite{8585034}. An obvious question is whether this new paradigm of computation can also be used for NLP. Such applicability may be to the end of leveraging the computational speedups for language-related problems, as well as for investigating how quantum systems, their mathematical description and the way information is encoded ``quantumly'' may lead to conceptual and practical advances in representing and processing language meaning beyond computational speedups.

Inspired by these prospects, \textit{Quantum Natural Language Processing} (QNLP), a field of research still in its infancy, aims at the development of NLP models explicitly designed to be executed on quantum hardware. There exists some impressive theoretical work in this area, but the proposed experiments are classically\footnote{In this paper the usage of the term `classical' is that of a qualifier in the sense of classical mechanics or classical probability theory, i.e. in contrast to quantum mechanics.}  simulated. A notable exception to this is a work by
\shortcite{meichanetzidis2020grammaraware}, where a proof of concept experiment with a dataset of 16 sentences was performed on quantum hardware for the first time.

In this paper we take a significant step further and present two \ed{complete 
experiments consisting of linguistically-motivated NLP tasks with datasets of the order of 100-150 sentences} running on quantum hardware. The goal of these experiments is not to demonstrate some form of ``quantum advantage'' over classical implementations of NLP tasks; for any practical application, this is not yet possible due to the limited capabilities of the currently available quantum computers. In this work, we are mostly interested in exploring the process of running NLP models on quantum hardware, and in providing a detailed account to the AI and NLP communities of what QNLP entails in practice. We show how the traditional modelling and coding paradigm can shift to a quantum-friendly form, and we take a closer look at the challenges and limitations imposed by the current \textit{Noisy Intermediate-Scale Quantum} (NISQ) computers.

From an NLP perspective, both of the tasks that this work considers involve some form of sentence classification: for each sentence in the dataset, we apply compositional models with various degrees of syntax sensitivity to compute a state vector, which is then converted to a binary label. The models are trained on a standard binary cross entropy objective, using an optimisation technique known as SPSA -- \textit{Simultaneous Perturbation Stochastic Approximation} \cite{Spall_1998_ImplementationGuideSPSA}. We implement quantum versions of the following models: 

\begin{itemize}
\item a bag-of-words model, where a sentence is represented as an unordered set of words, with no syntactic information present whatsoever.
\item a word-sequence model, in which the sentence is processed in a linear manner from left to right, as in a standard recurrent neural network. Since the model respects the order of the words, a limited degree of syntax is captured in this case.
\item a fully syntax-based model where composition takes place following a grammatical derivation 
given by a syntax tree. 
\end{itemize}

For the syntax-based case, which is admittedly the most complicated and interesting one from both a theoretical and an engineering point of view, we use the compositional model of \newcite{CoeckeSadrClark2010} -- often dubbed \discocat~(DIStributional COmpositional CATegorical). The choice of \discocat~is motivated by the fact that the derivations it produces essentially form a tensor network, which means they are already very close to how quantum computers process data. Furthermore, the model comes with a rigorous treatment of the interplay between syntax and semantics and with a convenient diagrammatic language. In Section \ref{sec:pipeline} we will see how the produced diagrams get a natural translation to quantum circuits -- the basic units of computation on a quantum computer\footnote{In the circuit-based, as opposed to the measurement-based, model of quantum computation.} -- and how sentences of different grammatical structure are mapped to different quantum circuits. 
We further discuss the role of noise on a NISQ device, and how this affects our design choices for the circuits. The experiments are performed on an IBM NISQ computer provided by the IBM Quantum platform\footnote{\url{https://quantum-computing.ibm.com}}.

For our experiments we use two different datasets. The first one (130 sentences) was generated automatically by a simple context-free grammar, with half of the sentences related to food and half related to IT (a binary classification task). The other one (105 noun phrases) was extracted from the \textsc{RelPron} dataset \shortcite{relpron}, and the goal of the model is to predict whether a noun phrase contains a subject-based or an object-based relative clause (again a binary classification task). As we will see later, although by any meaningful NLP standard the selected datasets and tasks are small-scale and rather trivial in nature, they already pose significant challenges for the current NISQ computers.

Despite the limitations, we demonstrate that all models converge smoothly, and that they produce good results (given the size of the datasets) in both simulated and quantum hardware runs. As a sanity check, a comparison of the performances of the various models reveals meaningful correlations between the degree of syntax sensitivity of the models and the tasks at hand: for example, the amount of syntax required to solve the meaning classification task is minimal, since a simple examination of each word in isolation is usually sufficient to produce the correct classification label for the sentence. This is however not true for the relative pronoun task, where syntactical structure starts to become important. The results we report follow and confirm these intuitions, showing that the bag-of-words model and the word-sequence model perform better on the former task, while \discocat~presents the best performance on the latter, as expected. 

In summary, the contributions of this paper are the following: 

\begin{itemize}
\item we propose quantum versions for a number of compositional NLP models;

\item we outline in some depth the process, the technicalities and the challenges of training and running an NLP model on a quantum computer;

\item we present the first set of meaningful NLP experiments on quantum hardware,  providing a strong proof of concept that quantum NLP is within our reach.
\end{itemize}

The structure of the paper is the following: Section \ref{sec:related_work} discusses the most important related work on experimental quantum computing, particularly with regard to Quantum Machine Learning (QML) and QNLP; 
Section \ref{sec:sentence_models} first briefly discusses NLP models for sentences in general and then describes in detail the \discocat\ model, as well as quantum-friendly models for sentence representation based on the \ws\ and \bow\ models; 
Section \ref{sec:Intro_QC} provides an introduction to quantum computing; Section \ref{sec:pipeline} gives a high-level overview for a general QNLP pipeline; Section \ref{sec:TheTasks} explains the tasks; Section \ref{sec:Experiments} \rl{provides all the necessary details for the experiments, and presents our results,} and finally, Section \ref{sec:Conclusion} summarises our findings and points to future work.  

\section{Related Work}
\label{sec:related_work}

On the matter of quantum computation in the NISQ era generally speaking, there is a plethora of hybrid classical-quantum algorithms with NISQ technology in mind \shortcite<for a review see>{bharti2021noisy,Preskill_2018_QuantumComputingInNISQEra}. 
These typically take the form of quantum machine learning (QML) protocols, the majority of which are based on \textit{variational quantum circuit} methods, where the parameters of a quantum circuit are trained through machine learning methods \shortcite<see, e.g.,>{Benedetti2019,JaderbergEtAl_2021_QuantumSelfSupervisedLearning,HarrowEtAl_2021_GradientsMeasurementsInHybridAlgorithms,HuangEtAl_2021_PowerOfDataInQML,FergusonEtAl_2021_MeasurementBasedVQE}.
However, useful quantum algorithms with theoretically proven speedups rely on fault-tolerant quantum computers, which are currently not available. (See Section~\ref{sec:Intro_QC} for more details on this aspect and basic concepts of quantum computation.)

Turning to works on quantum algorithms that specifically address language related problems (without employing the kind of model that this present work does) \shortciteA{bausch2021quantum} utilise Grover search to achieve superpolynomial speedups for the parsing problem,  
\shortciteA{wiebe2019quantum} use quantum annealing to solve problems for general context-free languages and  
\shortciteA{RAMESH2003103} provide quantum speedups for string-matching which is relevant for language recognition. 
Furthermore, in \shortcite{gallego2019language} parse trees are interpreted as an information-coarse-graining of tensor networks and it is also proposed that they can be instantiated as quantum circuits. 
To the best of our knowledge, any known quantum algorithm on language related problems with a proven speedup, including those listed above, is not implementable on today's NISQ technology.

Works that are not concerned with an implementation on quantum hardware, but study classical models for natural language that may be seen as  inspired by certain aspects of the mathematical formalism of quantum theory are abundant, but peacemeal. See, for instance, \shortcite{basile-tamburini-2017-towards} and \shortcite{ChenEtAl_2021_QuantumlanguageModel}. 
While not directly related to NLP, it is worth noting that there also is a lot of interesting work on quantum neural networks, see, for example \shortcite{GUPTA2001355,qnn}. 

Finally, the more specific context of this work, combining all of the above mentioned directions, is \emph{compositional QNLP} -- understood as approaches to a quantum implementation of NLP models in the spirit of the \textsc{DisCoCat} model. For an introduction to this model see Section~\ref{sec:discocat}. 
The early theoretical work by \shortciteA{Zeng2016} leverages the \textsc{DisCoCat} model to obtain a quadratic speedup for a sentence classification task.  
While, again, the proposed algorithm in that work requires technology that is currently not available, such as `quantum RAM', subsequent theoretical work \shortcite{coecke2020foundations} lays the foundation for implementations specifically on currently available NISQ devices. 
Furthermore, in \shortcite{O_Riordan_2020} a \textsc{DisCoCat}-inspired workflow is introduced along with experimental results obtained by classical simulation. 
\shortciteA{meichanetzidis2020grammaraware} provided for the first time a proof of concept that practical QNLP is in principle possible in the NISQ era, by managing to run and optimize on a NISQ device a classifier using a dataset of 16 artificial short sentences. This current work is a natural next step, presenting for the first time two NLP experiments of medium scale (given the quantum computing context) on quantum hardware.

\section{Compositional Models for Sentence Representation}
\label{sec:sentence_models}

\subsection{The General Landscape: From Syntax-Insensitive to Syntax-Sensitive Models} 
\label{sec:classical_model_overview}

In this section we briefly discuss the topic of sentence modelling and the role syntax plays in it. In the simplest case a sentence can be represented as a ``bag of words'', that is, an unordered set of symbols or most commonly \textit{embeddings}, that are associated with the words. Since each word is independent of its context, syntactic relationships cannot be modelled in this case. When embeddings or other distributional approaches are used, a simple compositional model for preparing a sentence representation would typically involve element-wise addition or multiplication of the vectors for the words in the sentence \cite{lapata2010}. For a few simple NLP tasks, where the examination of the words in isolation is enough to provide the correct result, a bag-of-words model can be a lightweight and effective solution.\footnote{The meaning classification task we describe in Section \ref{sec:TheTasks} is such an example.} 

However, for most practical real-world tasks, some amount of syntax is usually necessary. A \textit{word-sequence} model respects the order of the words, processing the sentence word by word from left to right. This approach can capture localised interactions between the words and (to some degree) longer-range dependencies as well. The most popular word-sequence model for NLP is the \textit{recurrent neural network} (RNN), with the \textit{long short-term memory} (LSTM) variation \cite{lstm} being the de-facto standard model for NLP-related tasks, at least before the advent of transformers. 

Finally, there is a class of models where the order of composition is strictly dictated by the syntactic structure of the sentence. 
For these cases, a parse tree is provided by a (typically statistical) parser, which for instance, for the sentence ``John gave Mary a flower'' would be of the following form:

\begin{figure}[H]
	\centering
	\ctikzfig{fig_parse_tree}
	\caption{Example of a parse tree with $S$ representing a sentence, $VP$ a verb phrase, $NP$ a noun phrase, $TV$ a transitive verb, $N$ a noun and $DT$ a determiner.}
\end{figure}

An example of a fully syntax-sensitive approach is the \textit{recursive neural network} \shortcite{socher2012}, in which a standard RNN composes the words of a sentence not in sequence from left to right, but by following a provided parse tree. These models have proven very effective in tasks such as sentiment analysis \shortcite{socher2013}. 

In light of the sketched spectrum of a model's syntax sensitivity, it is important to note that whether fully syntax-sensitive models are useful in NLP or not is still under debate. 
In fact, it has been found that in most cases modern neural network (NN) architectures are capable of dynamically learning the syntactic features that are required for the task at hand from the training data on-the-fly, making the need to explicitly provide syntax trees for the sentences obsolete \shortcite<see, e.g.,>{iyyer2015deep}. 
However, such a degree of learning capacity requires relatively large models and even larger amounts of training data, which is something that -- at least currently -- cannot be afforded in quantum computing given the status of the available machines. 
Thus, models intended to be run on quantum hardware have to be chosen wisely. 

The \discocat~model, which is introduced in the subsequent section (Sec.~\ref{sec:discocat}), is a fully syntax-sensitive model of natural language meaning and is particularly suitable for the purposes of this paper for two reasons. 
First, it features a particular structural compatibility with quantum theory. This will be explained in detail below, but in short the model's sentence representations can be seen as \textit{tensor networks} \cite{tensor_nets}. This makes the model a natural choice for a quantum implementation and more appropriate than above mentioned syntax-based models from conventional NLP. 
Second, to address the question of why use a syntax-based model in the first place, we emphasise again the tension between wanting a model that can deal with complex tasks that require syntax,
and the impossibility of putting on a quantum computer the kind of large model that learns aspects of syntax from the training data on-the-fly. 
Hence, the choice of \discocat~for the experiments of this work brings the additional benefit that a considerable amount of complexity is removed from the training process, since the syntactic structures of the sentences are provided to the model as part of the input.

\subsection{A Syntax-Based Model Inspired by Quantum Mechanics}
\label{sec:discocat}

This section introduces and explains in detail the \discocat~model of \newcite{CoeckeSadrClark2010}\footnote{The explanation of how one can implement this general, mathematical model for natural language meaning in different ways, on quantum or conventional/classical hardware, is postponed to Sections~\ref{sec:pipeline} and \ref{sec:Experiments}.}, whose aforementioned structural compatibility with quantum theory is due to the fact that it is based on the rigorous mathematical framework of compact closed categories. Compact closed categories are the structure that also provide an abstraction of the Hilbert space formulation of quantum theory \cite{abramsky2004}.     

In \discocat, the meaning of words is represented by tensors whose order is determined by the types of the words according to a \textit{pregroup grammar} \cite{lambek}. A type $p$ has a left ($p^l$) and a right adjoint ($p^r$), and the grammar has only two reduction rules:

\vspace{-0.3cm}
\begin{equation}
  p \cdot p^r \rightarrow 1 ~~~~~~~~~
  p^l \cdot p \rightarrow 1
\end{equation}
\vspace{-0.4cm}

Assuming atomic types $n$ for nouns and noun phrases and $s$ for sentences, the type of a transitive verb becomes $n^r \cdot s \cdot n^l$, denoting that an $n$ is expected on the left and another one on the right, to return an $s$. Thus, the derivation for a transitive sentence such as 
``John likes Mary'' is of the form:
\begin{multline}
  n \cdot (n^r \cdot s \cdot n^l) \cdot n \to 
  (n \cdot n^r) \cdot s \cdot (n^l \cdot n) \to 1 \cdot s \cdot 1 \to s
\end{multline}

\noindent
witnessing that it is a grammatical sentence. This derivation can also be represented diagrammatically as a \textit{pregroup diagram}:

\small
\ctikzfig{pregroup}
\normalsize

\noindent where the ``cups'' ($\cup$) denote the grammar reductions. The transition from pregroups to vector space semantics is achieved by a mapping\footnote{This mapping can be formalised as a category-theoretic \textit{functor} \cite{reasoning_2016}.} $\mathcal{F}$ that sends each atomic type (and its dual types) to a vector space ($n$ to $N$ and $s$ to $S$) and composite types to corresponding tensor product spaces ($n^r \cdot s \cdot n^l$ to $N\otimes S \otimes N$). For example, a transitive verb becomes a tensor of order 3, which can be seen as a bilinear map $N\otimes N \to S$, while an adjective (with type $n \cdot n^l$) is a matrix, i.e. a linear map $N \to N$. Further, $\mathcal{F}$ translates all grammar reductions to tensor contractions, so that the representation of a sentence $s = w_1w_2\dots w_n$ with $n$ words $w_1, w_2, \dots, w_n$ and a pregroup derivation $\alpha$ is given by:

\begin{equation}
  \mathbf{s} = \mathcal{F}(\alpha)\left[ \mathbf{w_1}\otimes \mathbf{w_2} \otimes \dots \otimes \mathbf{w_n} \right] 
  \label{equ:discocat}
\end{equation}

\noindent
where $\mathbf{w_i} = \mathcal{F}(w_i)$, i.e. $\mathbf{w_i}$ is the tensor that the $i$th word is mapped to, 
and $\mathcal{F}(\alpha)$, the image of the derivation $\alpha$ under $\mathcal{F}$, is a linear map that when applied to the tensor product of the word representations, by tensor-contracting that expression, returns a vector for the whole sentence.  
As a concrete example, the meaning of the sentence ``John likes Mary'' becomes

\ed{
\begin{equation}
\mathbf{s}_j \ =  \
\sum_{i,k} \mathbf{u}_i \ \mathbf{v}_{ijk} \ \mathbf{w}_k
\end{equation}
\noindent
where $\mathbf{u}, \mathbf{v} \in N$ are vectors -- tensors of order 1 -- representing the two nouns with $\mathbf{u}_i$ and $\mathbf{w}_k$ being their respective components with respect to a fixed basis. Similarly, $\mathbf{v} \in N\otimes S \otimes N$ is a tensor of order 3 representing the verb with components $\mathbf{v}_{ijk}$, and $\mathbf{s}$ is a vector in $S$ with components $\mathbf{s}_j$ representing the sentence .} 
Note that the underlying field of the vector spaces is not specified, and can, e.g., be $\mathbb{R}$ or $\mathbb{C}$ depending on the particular type of model. In this work the underlying field of the vector spaces is $\mathbb{C}$. 

Meaning computations like the example above can be conveniently represented using the diagrammatic calculus of compact closed categories \cite{Coecke_Kissinger_2017_TextBook}:

\small
\ctikzfig{diagram_adjusted}
\normalsize

\noindent where the boxes denote tensors, the order of which is determined by the number of their wires, and the cups are now the tensor contractions. Note the similarity between the pregroup diagram above with the one here, and how the grammatical derivation essentially dictates both the shapes of the tensors and the contractions. As we will see later, these \textit{string diagrams} can further naturally be mapped to quantum circuits for use in quantum computation.

There is a rich literature on the \discocat~model, and here we mention only a few indicative publications. Proposals for modelling verbs and implementations of the model have been provided by \newcite{grefenstette2011} and \shortcite{KartSadrPul-COLING-2013,kartsadrqpl2014}. \shortcite{SadrzadehEtAl_2013_FrobeniusAnatomyI} address the modelling of relative pronouns.  \shortcite{piedeleu2015} present a version of the model where the meaning of words is given by density matrices, encoding phenomena such as homonymy and polysemy. Various versions of the model have been used extensively in conceptual tasks such as textual entailment at the level of sentences, see for example \shortcite{SadrzadehEtAl_2018_SentenceEntailment,bankova,Lewis2019ModellingHF}. Finally, \newcite{yeung-kartsaklis} reformulate \discocat~as a passage from a Combinatory Categorial Grammar (CCG) \cite{syntactic-process} to a category of semantics, decoupling the model from pregroup grammars and making available to researchers a wide range of CCG resources, such as large-scale  collections of human-annotated syntactic trees and statistical parsers. 

\subsection{Quantum-Friendly Versions of the Word-Sequence and Bag-of-Words Models} 
\label{sec:basline_models}

As discussed, the \discocat~model is a syntax-based model that is suitable for quantum implementation. In order to enable a comparative study, and in light of the spectrum of syntax sensitivity sketched in Section~\ref{sec:classical_model_overview}, this section will propose and introduce examples of quantum-friendly versions of the \ws\ and \bow\ models.

Given the compatibility between tensor networks and the workings of a quantum computer a natural choice for a word-sequence model, simpler than the RNNs mentioned in Section~\ref{sec:classical_model_overview}, is simply to let a sentence representation be a tensor network that reflects the sentence's word order. 
For the purposes of this work, we propose the simple approach exemplified in Figure~\ref{fig:tensornet}, in which each word is a tensor of order 2 (a matrix) and their multiplication in the order as the respective words appear in the sentence, defines, once applied to a token ``start'' vector, the output vector that represents the overall sentence. 

\begin{figure}[!h]
\centering
\ctikzfig{fig_tensor_net}
\caption{A simple word-sequence model using a tensor network representation. The token $\langle S \rangle$ marks the start of a sentence.}
\label{fig:tensornet}
\end{figure}

It is instructive to note that this can equivalently be represented as a string diagram, as seen below in Figure~\ref{fig:WordSeqModel_as_StringDiagram}, by letting cups represent tensor contractions as before (see Section~\ref{sec:discocat}). The sentence representations in this model are therefore very similar those of \discocat. The differences are: a) that here the order of every word's tensor (its arity) is fixed to be the same independent from the words' grammatical types and b) that the connectivity of a diagram (how the tensors are contracted) is not determined by a parse-tree, but simply by the order in which the words appear in the sentence from left to right. 

\begin{figure}[H]
\centering
\ctikzfig{fig_tn_discocat_v2}
\caption{The diagram from Figure~\ref{fig:tensornet} rewritten in equivalent form as a string diagram.}
\label{fig:WordSeqModel_as_StringDiagram}
\end{figure}

Finally, in order to provide a suitable bag-of-words model, again making use of tensor network representations, we use a commutative linear operation that maps a number of vectors to another vector, where it is the commutativity that makes the insensitivity to the word order manifest. 
\ed{A suitable example of such an operation is provided by simple component-wise multiplication given a fixed basis $\{e_1,...,e_d\}$ of a $d$-dimensional vector space $V$. For $n$ input vectors, the map can be defined through its action on the fixed basis as follows
\begin{eqnarray}
	m \ : \ \bigotimes_{k=1}^n V &\rightarrow& V \nonumber \\
 			e_{i_1} \otimes \dots \otimes e_{i_n} &\mapsto& \delta_{i_1 \dots i_n} \ e_{i_1}  \nonumber
\end{eqnarray}
and, hence, some $n$ vectors, when represented in that fixed basis, are mapped to the vector that is obtained from component-wise multiplication: 
\[ \Big( \ \big(v_1^{(1)}, ..., v_d^{(1)} \big) \otimes ... \otimes \ \big(v_1^{(n)}, \ ..., v_d^{(n)} \big) \Big) 
\ \ \mapsto \ \ \Big( \ \big(\prod_{j=1}^n v_1^{(j)} \big), \ ..., \ \big(\prod_{j=1}^n v_d^{(j)} \big) \ \Big)
\]
This map has a common string diagrammatic representation as a `merge-dot' as in Figure~\ref{fig:discocat_spider},  seeing as it `merges' the $n$ copies of $V$ into a single copy of $V$\footnote{\ed{More abstractly, the map $m$ is an instance of the multiplication and co-multiplication maps, which are canonically induced by a fixed basis of a vector space, and form a \textit{Frobenius algebra}. See \newcite{reasoning_2016} for further explanation of how Frobenius multiplication and co-multiplication can be seen as copying and merging the dimensions of a tensor}}. 
In this simple model a sentence with $n$ words is thus represented by the vector obtained from applying the multiplication (merge operation) $m$ to the set of $n$ vectors that represent the words of that sentence. Figure~\ref{fig:discocat_spider} depicts the representation of our example sentence with 5 words in this model.} 

We note that once again the resulting diagrams such as the one in Figure~\ref{fig:discocat_spider}, are string diagrams and can be seen as even further simplified versions of \discocat\ representations. 

\begin{figure}[!h]
\centering
\ctikzfig{fig_spider_v2}
\caption{The example sentence in our bag-of-words model. The `dot' represents the \ed{multiplication  $m$, also called `merge' operation, defined above}.} 
\label{fig:discocat_spider}
\end{figure}

We will refer to these models simply as the \ws\ and \bow\ models, respectively. 
We emphasise again that they both yield sentence representations that are simpler versions of the same sort of objects as those of the \discocat~model, namely string diagrams that can be interpreted in terms of (complex) vector spaces and linear maps between them. 
This constitutes a compatibility between our models, which will be exploited in Sections~\ref{sec:pipeline} and \ref{sec:Experiments}, in order to apply a single unified experimental pipeline.

\section{Introduction to Quantum Computing}
\label{sec:Intro_QC}

For a serious introduction into quantum information theory and quantum computing, which is beyond the scope of this paper, the reader is referred to the literature \cite<see, e.g.,>{Coecke_Kissinger_2017_TextBook,NielsenChuang_2011_TextBook}. 
However, to provide a self-contained manuscript, in this section we will set out the required terms and concepts in such a way that no previous exposure to quantum theory is required.

\ed{We start with the concept of a \textit{qubit}, which, as the most basic unit of information carrier, is the quantum analogue of a bit, and yet a very different sort of thing.
It is associated with a property of a physical system such as the spin of an electron, which can be `up' or `down' along some axis and there is a sense in which a qubit can indeed be in two `extreme states', corresponding to the two possible outcomes of some measurement, 
hence the analogy to a classical bit. 
However, such pairs of extreme, namely perfectly distinguishable states do not exhaust the set of states a qubit can be in -- a general \textit{state} $\ket{\psi}$ lives in a 2-dimensional \textit{complex} vector space, more precisely a Hilbert space.\footnote{Such states $\ket{\psi}$ are also referred to as \textit{pure} states, as opposed to \textit{mixed} states. 
The latter are a different kind of object and in particular allow one to represent probabilistic mixtures over pure states when there is a lack of knowledge of which pure state was prepared. The formalism in this work will only make reference to pure states.} 
In denoting a state vector as $\ket{\psi}$, read `ket psi', we use notation that is common in physics and called \emph{braket} notation; why decorating a vector with a `ket' symbol, $\ket{ \ }$, is useful will become clear momentarily.
} 

\ed{
Let $\ket{0}$ and $\ket{1}$ denote orthonormal basis vectors of a qubit's Hilbert space, where the labels of these two states correspond to the respective outcomes of a measurement, which are here generically labelled as `0' and `1'.  
A general state of the qubit then is a (complex) linear combination known as a \textit{superposition}: 
$\ket{\psi} = \alpha \ket{0} + \beta \ket{1}$, where $\alpha, \beta \in \mathbb{C}$ and $|\alpha|^2 + |\beta|^2 = 1 $. 
A reader not familiar with complex valued linear algebra and Hilbert spaces will be relieved to learn that no such familiarity is necessary to follow the main content of this work, which will instead use a very intuitive diagrammatic language for these structures, introduced below. 
}

Importantly, quantum theory is a \textit{fundamentally probabilistic} theory, that is, even given that a qubit is in state $\ket{\psi}$ -- a state known as perfectly as is in principle possible -- this generally only allows one to make predictions for the probabilities, with which the outcomes `0' and `1', respectively, occur when the qubit is \textit{measured}.
\ed{These probabilities are given by the so-called Born rule $P(i) = |\braket{i|\psi}|^2$, where $i=0,1$ ranges over the possible outcomes\footnote{\rl{That is, the integer $i\in \{0,1\}$ is used to label vectors when writing $\ket{i}$ and $\bra{i}$.}} and 
$\braket{i|\psi}$ is a complex number, called the \emph{amplitude}, which is given by the inner product written as the composition of state $\ket{\psi}$ with \emph{bra} vector $\bra{i}$.
Hence, for the above state $\ket{\psi} = \alpha \ket{0} + \beta \ket{1}$ the outcome probabilities are given by $P(0)=|\alpha|^2$ and $P(1)= |\beta|^2$. 
}

\ed{We now also see the reason why braket notation is useful. 
The vector $\bra{i}$, which is decorated with a `bra', $\bra{ \ }$, denotes the dual vector of $\ket{i}$ in the corresponding dual Hilbert space (basically, $\ket{i}$ can be thought of as a column vector and $\bra{i}$ as the corresponding complex conjugated row vector) such that writing the \emph{bra-ket}, $\braket{i|\psi}$, gives the inner product of $\ket{\psi}$  and $\ket{i}$ -- the very thing that determines probability distributions as the empirical predictions of quantum theory. 
On a terminological note, a bra $\bra{i}$ is in the literature and in this manuscript also referred to as a (quantum) \emph{effect}, reflecting more intuitively that it corresponds to obtaining the measurement outcome $i$.}

\ed{According to quantum theory the \rl{evolution of an isolated qubit before measurement, that is the change of its state over time,} is described through the \textit{transformation} of its state with a \textit{unitary} linear map $U$,\footnote{\rl{A \emph{unitary} operator $U$ can be represented as a matrix $U$ that satisfies the condition of unitarity, i.e. $U^{\dagger}U = id = U U^{\dagger}$, where $id$ is the identity matrix on the vector space and $U^\dagger$ refers to the conjugate transpose of $U$.}} i.e. the state $\ket{\psi'}$ describing the closed system after such evolution relates to the initial state $\ket{\psi}$ as $\ket{\psi'} = U \ket{\psi}$. 
The amplitude for obtaining outcome $0$, given $\ket{\psi'}$ is $\braket{0|\psi'} = \bra{0} U \ket{\psi}$. 
The exact same linear equation is represented in Fig.~\ref{fig:Example_BasicSeqComp}, now starting to introduce the intuitive diagrammatic language for quantum theory and quantum computing promised earlier: 
the diagrams are read top down; 
the box labelled $U$ represents the unitary linear map $U$, while the input and output `wires' of the box correspond to qubits; 
triangles without input wires represent states, while triangles that are orientated upside-down compared to those of states and without output wires represent effects -- the (non-deterministic) outcomes of measurements.
}

\ed{More generally, the} joint state space of $q$ qubits is given by the tensor product of their individual state spaces and thus has (complex) dimension $2^q$. 
For instance, for two `uncorrelated' qubits in states $\ket{\psi_1} = \alpha_1 \ket{0} + \beta_1 \ket{1}$ and $\ket{\psi_2} = \alpha_2 \ket{0} + \beta_2 \ket{1}$, the joint state is $\ket{\psi_1} \otimes \ket{\psi_2}$, which in basis-dependent notation becomes $(\alpha_1, \beta_1)^T \otimes (\alpha_2, \beta_2)^T = (\alpha_1 \alpha_2, \ \alpha_1\beta_2, \ \beta_1\alpha_2, \ \beta_1\beta_2)^T$. 
The evolution of a set of qubits that interact with each other, is then described by a unitary map acting on the overall state space. 

\ed{The diagrammatic representation from Fig.~\ref{fig:Example_BasicSeqComp} extends naturally to \textit{quantum circuits} for the general case of several qubits \rl{\cite<see, e.g.,>{Coecke_Kissinger_2017_TextBook,Coecke_2022_QuantumInPictures}} -- the different qubits are represented as parallel wires, and their evolution may now involve boxes with more than one input and output wire seeing as the evolution will generally involve interaction amongst the qubits. 
An example is shown in Fig.~\ref{fig:Example_circuit}, where 5 qubits are prepared in an initially uncorrelated state of all $\ket{0}$ states and then evolve through a sequence of unitaries.  
The overall diagram represents the output state, i.e. the state after the evolution happened, at which point the qubits may be measured or may keep evolving further.
}

\begin{figure}
	\centering
	\begin{subfigure}{0.15\textwidth}
		\centering
		\small
		\ctikzfig{Example_BasicSeqComp}
	    \normalsize
		\caption{\label{fig:Example_BasicSeqComp}}
	\end{subfigure}
	\hspace*{1.5cm}
	\begin{subfigure}{0.3\textwidth}
		\small		
		\ctikzfig{figure01b_corrected}
		\caption{\label{fig:Example_circuit}}
	\end{subfigure}
	\caption{Basic examples of quantum circuits as a diagrammatic language for quantum theory and quantum computing (see main text for details).
	}
\end{figure}

The example quantum circuit in Fig.~\ref{fig:Example_circuit} contains all the kinds of gates \rl{that make appearance in this paper:} 
the Hadamard gate $H$ is a specific single qubit unitary transformation; similarly, the quantum CNOT gate in $(ii)$ is a specific two-qubit unitary transformation;  
the gate $Rx(\beta)$, denotes a parameterised unitary, namely for every $\beta \in [0, 2 \pi]$ it is an $X$-rotation by angle $\beta$; 
and finally, there is the controlled $Z$-rotation gate in $(i)$, a component of which is a $Z$-rotation gate $Rz(\delta)$ by angle $\delta \in [0, 2 \pi]$. 
\rl{Note that knowing the precise definitions of these gates as specific linear maps is irrelevant to understanding the rest of this work, all one needs to know is that the symbols $H$, $Rx(\beta)$ etc. refer to the sorts of maps as just described. For the sake of completeness, however,  Fig.~\ref{Fig_Def_Basic_gates} gives their definitions in terms of what the respective linear map does to each basis vector.}

\begin{figure}
	\centering
	\begin{minipage}{\textwidth}
		\centering
		\begin{minipage}{0.35\textwidth}
			\begin{eqnarray}
				H \ : \  \ket{0} &\mapsto& \frac{1}{\sqrt{2}} \big( \ket{0} + \ket{1} \big) \nonumber \\
				 \ket{1} &\mapsto& \frac{1}{\sqrt{2}} \big( \ket{0} - \ket{1} \big) \nonumber \\[0.5cm]
				 Rz(\alpha) \ : \  \ket{0} &\mapsto& e^{-i\frac{\alpha}{2}} \ket{0} \nonumber \\
				\ket{1} &\mapsto& e^{i\frac{\alpha}{2}} \ket{1} \nonumber
			\end{eqnarray}
		\end{minipage}
		\hfill
		\begin{minipage}{0.6\textwidth}
			\begin{eqnarray}
				Rx(\beta) \ : \  \ket{0} &\mapsto& \text{cos}(\frac{\beta}{2})\ket{0} - i \text{sin}(\frac{\beta}{2})\ket{1} \nonumber \\
				  \ket{1} &\mapsto& - i \text{sin}(\frac{\beta}{2})\ket{0} + \text{cos}(\frac{\beta}{2})\ket{1} \nonumber \\[0.5cm]
				 CNOT \ : \  \ket{00} &\mapsto& \ket{00} \ , \ \ \ket{10} \ \mapsto \ \ket{11} \nonumber \\
				\ket{01} &\mapsto& \ket{01} \ , \ \ \ket{11} \ \mapsto \ \ket{10} \nonumber 
			\end{eqnarray}
		\end{minipage}
	\end{minipage} 
	\caption{\ed{Definition of some basic quantum gates with $\alpha, \beta \in [0, 2 \pi]$. The controlled $Z$-rotation gate in $(i)$ of Fig.~\protect\ref{fig:Example_circuit} is defined analogously to how the quantum CNOT gate features `controlling', i.e. for the control qubit in $\ket{0}$ the target qubit is unchanged, whereas for $\ket{1}$ a $Z$-rotation is applied to the target qubit.} \label{Fig_Def_Basic_gates}}
\end{figure}

\ed{So on the one hand, a quantum circuit captures the structure of the overall linear map evolving the respective qubits, where parallel `triangles' and parallel boxes are to be read as tensor products of states and unitary maps, respectively, and sequential `wiring up' of boxes as composition of linear maps. 
Hence, a circuit as a whole represents the application of a linear map to a vector, computing the systems' overall state at a later time -- in other words, it is simple linear algebra and can be viewed as tensor contraction of complex valued tensors that are represented by the gates. 
On the other hand,} a circuit can conveniently also be seen as a list of abstract instructions for what to actually \textit{implement} on some quantum hardware in order to make, say, some ions\footnote{The basic physical object used as a qubit varies vastly across different quantum computers.} undergo, physically, the corresponding transformations as prescribed by the gates of the circuit. 

Now, coming back to the fact that quantum theory is probabilistic, once a circuit has been run on hardware all qubits are measured. In the case of Fig.~\ref{fig:Example_circuit} this yields 5 bits each time and many such runs have to be repeated to obtain statistics from which to estimate the outcome probabilities. These probabilities connect theory and experiment.
In order to obtain the result for a given problem, the design of the circuit has to encode the problem such that that result is a function of the outcome probabilities. Hence, the choice of circuit is key. 

A special case worth mentioning due to its relevance for this paper is as follows:
the encoding of the quantity of interest in a circuit over, say, $q$ qubits may be such that the result is a function of the outcome distribution on just $r$ of the qubits ($r < q$), but subject to the condition that the remaining $q-r$ qubits have yielded particular outcomes, i.e. the result is a function of the corresponding conditional probability distribution.  
The technical term for this is \textit{post-selection}, as one has to run the whole circuit many times and measure all qubits to then restrict -- post-select -- the data for when the condition on the $q-r$ qubits is satisfied. 
At the diagrammatic level the need for such post-selection is typically indicated by the corresponding quantum effects as done, e.g., in Fig.~\ref{fig:ExampleSent_QCircuit}, which up to the 0-effects on 4 of the 5 qubits, is identical to that of Fig.~\ref{fig:Example_circuit}. 


Finally and crucially, actually building and running a quantum computer is
a challenging engineering task for multiple reasons. 
Above all, qubits are prone to random errors from their environment and unwanted interactions amongst them. 
This `coherent noise' is different in nature to that of a classical computing hardware. 
A quantum computer that would give the expected advantages for large scale problems, is one that comes with a large number of \textit{fault-tolerant} qubits, essentially obtained by clever error correction techniques.
Quantum error correction \shortcite{Brown_2016} reduces to finding ways for distributively encoding the state of a logical qubit on a large number of physical qubits (hundreds or even thousands).
The scalable technical realisability of logical qubits is still out of reach at the time of writing. 
The currently available quantum devices are rather noisy medium-scale machines with qubit numbers 
\rl{mostly still in the double digit range.} 
These devices provide proof of concept and are extremely valuable assets for the development of both theory and applications.
This is the reason one speaks of the NISQ era, mentioned in Sec.~\ref{sec:intro}, and this is the light in which the work in this paper has to be seen -- exciting proof of concept, while the machines are still too small and noisy for large-scale QNLP experiments.

\section{The General Pipeline}
\label{sec:pipeline}

In Section~\ref{sec:Intro_QC} we saw that the quantum computing equivalent to classical programming involves the application of quantum gates on qubits according to a quantum circuit. 
This section will explain the general pipeline of our approach, and in particular the process of how to go from a sentence to its representation as a quantum circuit, which forms the basis on which the model predicts the label. 
We first do this explicitly for the \discocat~model as it constitutes the most intricate of the three models discussed in Section~\ref{sec:sentence_models}. We then make clear how the other two models are subsumed in this pipeline by simply dropping or simplifying particular steps.  

\begin{figure}
\begin{center}
\includegraphics[scale=0.5]{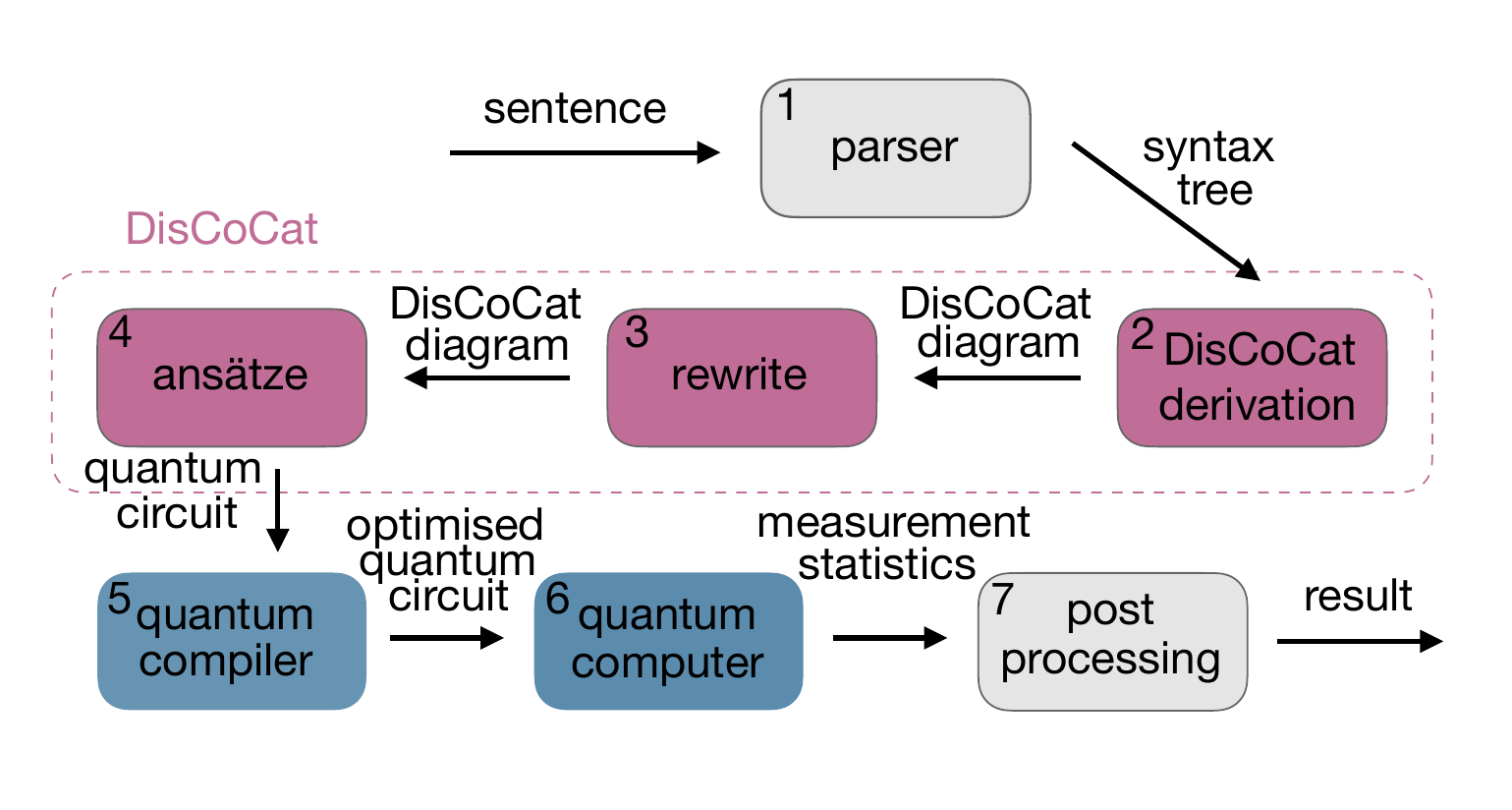}
\end{center}
\caption{Schematic overview of the general pipeline.}
\label{fig:pipeline}
\end{figure}

Figure~\ref{fig:pipeline} schematically depicts this pipeline and in this section we address each numbered step at a generic level. Concrete examples of the choices that one has to make in each step are covered in the implementation of this pipeline presented in Section~\ref{sec:Experiments}. 

Since \discocat~is fully syntax-sensitive, the first step is to get a syntax tree corresponding to the sentence. From this a \discocat~derivation is created in a diagrammatic form. 
In order to avoid computational complications on quantum hardware, this diagram is first optimised to yield the input into an \textit{ansatz} which determines the actual conversion into a quantum circuit. 
A quantum compiler translates the quantum circuit into hardware-specific code that can be run on quantum hardware. 
These stages are described in more detail below. 

\textit{Step 1 (Parser):} 
For a large-scale NLP experiment with thousands of sentences of various structures, the use of a pregroup parser for providing the syntax trees would of course be necessary.\footnote{Since to the best of our knowledge at the time of writing there are not any robust pregroup parsers available, an alternative approach would be to use a CCG parser and subsequently convert the types into pregroups. \newcite{yeung-kartsaklis} provide the details of a functorial passage from CCG derivations encoded in a biclosed category to \discocat~diagrams, and a web tool that demonstrates the conversion  (\url{https://qnlp.cambridgequantum.com/generate.html}).
Furthermore, the open-source Python package \texttt{lambeq} \shortcite{kartsaklis2021lambeq} fully automates the conversion of a sentence to a monoidal diagram and then to a quantum circuit by coupling to a CCG parser and enables the user to design and implement compositional QNLP experiments; more information can be found at \url{https://cqcl.github.io/lambeq}.}
In the present work this step can however be executed semi-automatically due to the limited vocabulary and the small number of different grammatical structures in our sentences, \ed{that is, once one has worked out the grammatical derivations for all types of sentences or phrases appearing in a dataset, a simple look-up table on the basis of the types of words appearing in the sentence or phrase, allows one to produce the respective parsing.}  
For instance, with nouns, adjectives and transitive verbs having the types $n$, $n \cdot n^l$ and $n^r \cdot n \cdot n^l$,  respectively, the sentence ``person prepares tasty dinner" is parsed as below \ed{(see Sec. \ref{sec:discocat} for more details on the pregroup grammar and Sec.~\ref{sec:Experiments} for details on the specific datasets studied in this work)}:  
\begin{multline}
   n \cdot (n^r \cdot s \cdot n^l) \cdot (n \cdot n^l) \cdot n  \to  
   (n \cdot n^r) \cdot s \cdot (n^l \cdot n) \cdot (n^l \cdot n) \to  
   1 \cdot s \cdot 1 \cdot 1 \to s
\end{multline}

\textit{Step 2 (\textsc{\textit{DisCoCat}} Derivation):}  Construct the sentences' \discocat~diagrams by representing each word as a state, i.e. a box, and then `wiring them up' by drawing a cup for every reduction rule. The above example becomes:\footnote{As discussed in Sec. \ref{sec:discocat}, the diagram in Fig.~\ref{fig:ExampleSent_PGDiagram} represents linear-algebraic operations between tensors in vector spaces $N$, $S$ and tensor products of them. For convenience, we also include the pregroup types as a reminder of the grammatical rules involved in each contraction. For the remainder of this paper, only the pregroup types will be shown since they are indicative of the vector spaces they are mapped to.}

\begin{figure}[H]
\centering

\begin{tikzpicture}
	\begin{pgfonlayer}{nodelayer}
		\node [style=none] (0) at (-9.75, 6.25) {};
		\node [style=none] (1) at (-10.5, 4.5) {};
		\node [style=none] (2) at (-6, 4.5) {};
		\node [style=none] (3) at (-6.75, 6.25) {};
		\node [style=none] (4) at (-8.25, 5.25) {\small{person}};
		\node [style=none] (5) at (-4.5, 6.25) {};
		\node [style=none] (6) at (-5.25, 4.5) {};
		\node [style=none] (7) at (0.25, 4.5) {};
		\node [style=none] (8) at (-0.5, 6.25) {};
		\node [style=none] (9) at (-2.5, 5.25) {\small{prepares}};
		\node [style=none] (10) at (1.75, 6.25) {};
		\node [style=none] (11) at (1, 4.5) {};
		\node [style=none] (12) at (5.5, 4.5) {};
		\node [style=none] (13) at (4.75, 6.25) {};
		\node [style=none] (14) at (3.25, 5.25) {\small{tasty}};
		\node [style=none] (15) at (7, 6.25) {};
		\node [style=none] (16) at (6.25, 4.5) {};
		\node [style=none] (17) at (10.75, 4.5) {};
		\node [style=none] (18) at (10, 6.25) {};
		\node [style=none] (19) at (8.5, 5.25) {\small{dinner}};
		\node [style=none] (20) at (-8.25, 2.25) {};
		\node [style=none] (21) at (-4.25, 2.25) {};
		\node [style=none] (22) at (-2.5, 2.25) {};
		\node [style=none] (23) at (-0.75, 2.25) {};
		\node [style=none] (24) at (2, 2.25) {};
		\node [style=none] (25) at (4.5, 2.25) {};
		\node [style=none] (26) at (8.5, 2.25) {};
		\node [style=none] (27) at (-2.5, 0.75) {};
		\node [style=none] (30) at (-1.25, 1) {\small{$n^l$}};
		\node [style=none] (31) at (2.25, 1.25) {\small{$n$}};
		\node [style=none] (32) at (5.5, 1.75) {\small{$n^l$}};
		\node [style=none] (33) at (9, 1.75) {\small{$n$}};
		\node [style=none] (34) at (-8.75, 1.75) {\small{$n$}};
		\node [style=none] (35) at (-5.25, 1.75) {\small{$n^r$}};
		\node [style=none] (37) at (-3.25, 1.25) {\small{$s$}};
		\node [style=none] (38) at (-8.25, 4.5) {};
		\node [style=none] (39) at (-8.25, 3.75) {};
		\node [style=none] (40) at (-4.25, 3.75) {};
		\node [style=none] (41) at (-2.5, 3.75) {};
		\node [style=none] (42) at (-0.75, 3.75) {};
		\node [style=none] (43) at (-4.25, 4.5) {};
		\node [style=none] (44) at (-2.5, 4.5) {};
		\node [style=none] (45) at (-0.75, 4.5) {};
		\node [style=none] (46) at (2, 4.5) {};
		\node [style=none] (47) at (2, 3.75) {};
		\node [style=none] (48) at (8.5, 4.5) {};
		\node [style=none] (49) at (8.5, 3.75) {};
		\node [style=none] (50) at (4.5, 4.5) {};
		\node [style=none] (51) at (4.5, 3.75) {};
		\node [style=none] (52) at (-8.25, 3) {\small{N}};
		\node [style=none] (53) at (-4.25, 3) {\small{N}};
		\node [style=none] (54) at (-2.5, 3) {\small{S}};
		\node [style=none] (59) at (8.5, 3) {\small{N}};
		\node [style=none] (60) at (4.5, 3) {\small{N}};
		\node [style=none] (61) at (2, 3) {\small{N}};
		\node [style=none] (62) at (-0.75, 3) {\small{N}};
	\end{pgfonlayer}
	\begin{pgfonlayer}{edgelayer}
		\draw (1.center) to (0.center);
		\draw (0.center) to (3.center);
		\draw (3.center) to (2.center);
		\draw (1.center) to (2.center);
		\draw (6.center) to (5.center);
		\draw (5.center) to (8.center);
		\draw (8.center) to (7.center);
		\draw (6.center) to (7.center);
		\draw (11.center) to (10.center);
		\draw (10.center) to (13.center);
		\draw (13.center) to (12.center);
		\draw (11.center) to (12.center);
		\draw (16.center) to (15.center);
		\draw (15.center) to (18.center);
		\draw (18.center) to (17.center);
		\draw (16.center) to (17.center);
		\draw [bend right=90, looseness=1.25] (20.center) to (21.center);
		\draw (22.center) to (27.center);
		\draw [bend right=90, looseness=1.25] (23.center) to (24.center);
		\draw [bend right=90, looseness=1.25] (25.center) to (26.center);
		\draw (38.center) to (39.center);
		\draw (43.center) to (40.center);
		\draw (44.center) to (41.center);
		\draw (45.center) to (42.center);
		\draw (46.center) to (47.center);
		\draw (48.center) to (49.center);
		\draw (50.center) to (51.center);
	\end{pgfonlayer}
\end{tikzpicture}}

\caption{ }
\label{fig:ExampleSent_PGDiagram}
\end{figure}

\textit{Step 3 (Rewrite):} 
The structure of compact closed categories comes with rewrite rules that allow the transformation of diagrams such as the one shown above into equivalent ones. 
\ed{The reason this is relevant is that a different and yet equivalent string diagram can come with computational advantages when actually running a model. 
\rl{These advantages can be hardware specific -- e.g. that a particular gate is more costly to implement than others, or hardware agnostic and due to general facts about quantum information processing.}
Ideally, one would of course want a comprehensive optimisation algorithm that, when given a model, a task and a chosen quantum hardware finds the most efficient string diagram. However, such does not exist yet and, in any case given how recent the field is, it is too early to tell what the kinds of hardware and models will be in future that the algorithm should range over.}

\ed{The purpose of this exposition here is to make two points. 
First, for any given model and experimental implementation, the pipeline we are presenting will typically involve some rewriting at this stage for said reasons. 
Second, for the particular models this work studies there is indeed an obvious and very useful rewrite procedure. We will make a start with explaining it here, as this is where in the pipeline it is taking place, while a full understanding of why this method makes sense will built up through the discussion of the remainder of the pipeline below.}

\ed{Our specific rewrite method has to do with two facts: cups are costly to implement -- basically because they involve obtaining particular measurement outcomes on several qubits that only occur non-deterministically (see Steps 6 and 7 for more details) -- but there is a simple, and yet effective equivalence transformation on our string diagrams that reduces the number of cups. The latter is achieved by `bending down' all nouns of a sentence, by which we mean that, for instance, the string diagram from Fig.~\ref{fig:ExampleSent_PGDiagram} is mapped to the following diagram:}

\begin{figure}[H]
\centering

\begin{tikzpicture}
	\begin{pgfonlayer}{nodelayer}
		\node [style=none] (0) at (-7.75, 0.5) {};
		\node [style=none] (1) at (-8.5, 2) {};
		\node [style=none] (2) at (-3.5, 2) {};
		\node [style=none] (3) at (-4.25, 0.5) {};
		\node [style=none] (4) at (-6, 1.25) {\small{person}};
		\node [style=none] (5) at (-4.75, 6) {};
		\node [style=none] (6) at (-5.5, 4.5) {};
		\node [style=none] (7) at (0.25, 4.5) {};
		\node [style=none] (8) at (-0.5, 6) {};
		\node [style=none] (9) at (-2.5, 5.25) {\small{prepares}};
		\node [style=none] (10) at (1.5, 6) {};
		\node [style=none] (11) at (0.75, 4.5) {};
		\node [style=none] (12) at (4.75, 4.5) {};
		\node [style=none] (13) at (4, 6) {};
		\node [style=none] (14) at (2.75, 5.25) {\small{tasty}};
		\node [style=none] (15) at (4, 0.5) {};
		\node [style=none] (16) at (3.25, 2) {};
		\node [style=none] (17) at (7.75, 2) {};
		\node [style=none] (18) at (7, 0.5) {};
		\node [style=none] (19) at (5.5, 1.25) {\small{dinner}};
		\node [style=none] (20) at (-6, 2) {};
		\node [style=none] (21) at (-3.5, 4.5) {};
		\node [style=none] (22) at (-2.25, 4.5) {};
		\node [style=none] (23) at (-1, 4.5) {};
		\node [style=none] (24) at (2, 4.5) {};
		\node [style=none] (25) at (3.5, 4.5) {};
		\node [style=none] (26) at (5.5, 2) {};
		\node [style=none] (27) at (-2.25, 0) {};
		\node [style=none] (30) at (0, 3.75) {\small{$n^l$}};
		\node [style=none] (31) at (1.25, 3.75) {\small{$n$}};
		\node [style=none] (32) at (4.75, 3.75) {\small{$n^l$}};
		\node [style=none] (35) at (-4.5, 3.75) {\small{$n^r$}};
		\node [style=none] (37) at (-2.75, 1) {\small{$s$}};
		\node [style=none] (38) at (-4.5, 3.25) {};
		\node [style=none] (39) at (4.5, 3.25) {};
	\end{pgfonlayer}
	\begin{pgfonlayer}{edgelayer}
		\draw (1.center) to (0.center);
		\draw (0.center) to (3.center);
		\draw (3.center) to (2.center);
		\draw (1.center) to (2.center);
		\draw (6.center) to (5.center);
		\draw (5.center) to (8.center);
		\draw (8.center) to (7.center);
		\draw (6.center) to (7.center);
		\draw (11.center) to (10.center);
		\draw (10.center) to (13.center);
		\draw (13.center) to (12.center);
		\draw (11.center) to (12.center);
		\draw (16.center) to (15.center);
		\draw (15.center) to (18.center);
		\draw (18.center) to (17.center);
		\draw (16.center) to (17.center);
		\draw (22.center) to (27.center);
		\draw [bend right=90, looseness=2.00] (23.center) to (24.center);
		\draw [bend left=45, looseness=1.25] (21.center) to (38.center);
		\draw [bend right] (38.center) to (20.center);
		\draw [bend right=45, looseness=1.25] (25.center) to (39.center);
		\draw [bend left=45, looseness=1.25] (39.center) to (26.center);
	\end{pgfonlayer}
\end{tikzpicture}}

\caption{ }
\label{fig:ExampleSent_PGDiagram_bent}
\end{figure} 


Here the states for `person' and `dinner' have been `bent' down, which reduced the overall number of cups in the string diagram from three to a single one. 
That the diagrams really are representing the exactly same mathematical object is basically down to the structure of \emph{compact closure}. 
\rl{While the formal details behind it can be read up, e.g., in \cite{Coecke_Kissinger_2017_TextBook}, the relevant core fact here is intuitive and easy to understand. 
Compact closure has to do with the presence of the cups, which at the level of the pregroup grammar represent type reductions (see Sec.~\ref{sec:discocat}). 
The above equivalence transformation on the diagram bends a word box down and represents it with a box that -- recalling diagrams are read top down -- has an ingoing wire but no outgoing one. 
This is in contrast to the respective original word boxes from Fig.~\ref{fig:ExampleSent_PGDiagram}, which have no ingoing wire, but an outgoing one. 
That such rewriting is allowed is due to a 1-to-1 correspondence between these two representations of words. 
This correspondence can be seen to be established by simply defining the corresponding `turned around' box as the composition of the original noun with a cup as follows:}
\begin{equation}
	
\begin{tikzpicture}
	\begin{pgfonlayer}{nodelayer}
		\node [style=none] (0) at (-19.5, 0) {};
		\node [style=none] (1) at (-20.25, 1.5) {};
		\node [style=none] (2) at (-15.25, 1.5) {};
		\node [style=none] (3) at (-16, 0) {};
		\node [style=none] (4) at (-17.75, 0.75) {\small{person}};
		\node [style=none] (15) at (4.5, 0) {};
		\node [style=none] (16) at (3.75, 1.5) {};
		\node [style=none] (17) at (8.25, 1.5) {};
		\node [style=none] (18) at (7.5, 0) {};
		\node [style=none] (19) at (6, 0.75) {\small{dinner}};
		\node [style=none] (20) at (-17.75, 1.5) {};
		\node [style=none] (26) at (6, 1.5) {};
		\node [style=none] (32) at (5, 3.5) {\small{$n^l$}};
		\node [style=none] (35) at (-18.75, 3.5) {\small{$n^r$}};
		\node [style=none] (38) at (-17.75, 3.5) {};
		\node [style=none] (39) at (6, 3.5) {};
		\node [style=none] (40) at (-11.5, 1.5) {};
		\node [style=none] (41) at (-12, 0) {};
		\node [style=none] (42) at (-7.5, 0) {};
		\node [style=none] (43) at (-8, 1.5) {};
		\node [style=none] (44) at (-9.75, 0.75) {\small{person}};
		\node [style=none] (55) at (14.5, 1.5) {};
		\node [style=none] (56) at (14, 0) {};
		\node [style=none] (57) at (18.5, 0) {};
		\node [style=none] (58) at (18, 1.5) {};
		\node [style=none] (59) at (16.25, 0.75) {\small{dinner}};
		\node [style=none] (60) at (-9.75, -0.5) {};
		\node [style=none] (61) at (-5.75, -0.5) {};
		\node [style=none] (62) at (-5.75, 3.5) {};
		\node [style=none] (65) at (12.25, -0.25) {};
		\node [style=none] (66) at (16.25, -0.25) {};
		\node [style=none] (67) at (-5.75, -0.5) {};
		\node [style=none] (70) at (13.25, 3.5) {\small{$n^l$}};
		\node [style=none] (73) at (-6.75, 3.5) {\small{$n^r$}};
		\node [style=none] (85) at (12.25, 3.5) {};
		\node [style=none] (86) at (12.25, -0.25) {};
		\node [style=none] (87) at (-13.5, 1) {$:=$};
		\node [style=none] (88) at (10, 1) {$:=$};
		\node [style=none] (89) at (-10.5, -0.75) {\small{$n$}};
		\node [style=none] (90) at (17, -0.75) {\small{$n$}};
		\node [style=none] (91) at (16.25, 0) {};
		\node [style=none] (92) at (16.25, -0.25) {};
		\node [style=none] (93) at (-9.75, 0) {};
		\node [style=none] (94) at (-9.75, -0.5) {};
	\end{pgfonlayer}
	\begin{pgfonlayer}{edgelayer}
		\draw (1.center) to (0.center);
		\draw (0.center) to (3.center);
		\draw (3.center) to (2.center);
		\draw (1.center) to (2.center);
		\draw (16.center) to (15.center);
		\draw (15.center) to (18.center);
		\draw (18.center) to (17.center);
		\draw (16.center) to (17.center);
		\draw (38.center) to (20.center);
		\draw (39.center) to (26.center);
		\draw (41.center) to (40.center);
		\draw (40.center) to (43.center);
		\draw (43.center) to (42.center);
		\draw (41.center) to (42.center);
		\draw (56.center) to (55.center);
		\draw (55.center) to (58.center);
		\draw (58.center) to (57.center);
		\draw (56.center) to (57.center);
		\draw [in=270, out=-90, looseness=1.25] (60.center) to (61.center);
		\draw (62.center) to (67.center);
		\draw [in=-90, out=-90, looseness=1.50] (65.center) to (66.center);
		\draw (85.center) to (86.center);
		\draw (91.center) to (92.center);
		\draw (93.center) to (94.center);
	\end{pgfonlayer}
\end{tikzpicture}}

	\nonumber
\end{equation}

\rl{This correspondence does not only exist at the level of the pregroup grammar, but also at the level of the (complex) vector spaces that the pregroup types then get mapped to -- here this correspondence can be thought of as the isomorphism between column vectors and row vectors through the operation of taking the (conjugate) transpose.}

Above rewrite procedure thus generally reduces the number of cups by as many nouns as are present in the sentence.

\textit{Step 4 (Ans{\"a}tze):} \ed{In this step a sentence's abstract \discocat~representation takes a more concrete form; its \discocat~diagram is mapped to a specific parametrised quantum circuit.} 
\ed{This map is determined by choosing: 
\begin{enumerate}[label=(\alph*)]
	\item the number $q_n$ and $q_s$ of qubits that every wire of type $n$ and $s$, respectively, as well as their dual types, get mapped to; 
	\item concrete parametrised quantum states (effects) that all word states (effects) get consistently replaced with. 
\end{enumerate}} 
We refer to the conjunction of such choices as an \ed{\textit{ansatz}. Principled approaches to choosing (b) are presented in Sec.~\ref{sec:Experiments}, but for an illustration consider again the example from Fig.~\ref{fig:ExampleSent_PGDiagram_bent} translated into a parametrised} quantum circuit of the form as shown in Fig.~\ref{fig:ExampleSent_QCircuit}. 

\begin{figure}[h]
\centering

\InputIfFileExists{figure05_corrected.tikz}{}{\input{.//tikz_figures//figure05_corrected.tikz}}

\caption{Example of interpreting Fig.~\ref{fig:ExampleSent_PGDiagram_bent} as a quantum circuit according to an ansatz. 
\ed{Here $q_n=1=q_s$, i.e. one qubit per sentence and noun type; 
the words `prepares' and `tasty' are replaced with the parametrised quantum states as marked by (i) and (ii), respectively; 
the words `person' and `dinner' are replaced with the parametrised quantum effects as marked by (iii) and (v), respectively; 
the component (iv) is a concrete representation of the cup in quantum circuits (NB an unparametrised quantum effect).}}
\label{fig:ExampleSent_QCircuit}
\end{figure} 

\ed{Once an ansatz is chosen such as the one sketched in Fig.~\ref{fig:ExampleSent_QCircuit}, a concrete embedding of some word is thus fixed by further fixing particular values for the parameters that appear in the respective word's parametrised quantum state (effect). 
For instance, the word `tasty' from Fig.~\ref{fig:ExampleSent_PGDiagram_bent} is mapped to the component (ii) of Fig.~\ref{fig:ExampleSent_QCircuit}, which for every $\delta \in [0,2 \pi]$ prepares a specific two-qubit quantum state. 
Note that under this functorial mapping according to some ansatz any cup\footnote{Recall from the discussion in Sec. \ref{sec:discocat} that cups can also be seen to correspond to tensor contractions.} in a \discocat~diagram has an already fixed meaning, it is a particular quantum effect,\footnote{\ed{In the physics literature also known as a \textit{Bell-effect}.}} which can be represented as done in the component (iv) of Fig.~\ref{fig:ExampleSent_QCircuit}. 
}

It is worth emphasising that \ed{the output of the mapping is a parametrised quantum circuit} whose connectivity is fixed by the sentence's syntax, while the choice of ansatz determines the number of parameters for each word's representation. 

\ed{Now, in principle it is of course known how many parameters $p$ are needed to fix the most general state on $q$ qubits, so why does this choice of ansatz matter} (independent from questions of overfitting and generalisation)? There are two reasons, which are both of a practical nature. 
First, $p$ is exponential in $q$. So, even beyond the NISQ era, for the sort of dataset sizes and lengths of sentences one wishes to consider in NLP, a feasible number of parameters has to be achieved. \ed{This is to say, in practice one will rarely afford to work with a fully general parametrised quantum state that can `hit' any state in some multi-qubit state space for some choice of the parameters.} 
Second, different quantum machines have different sets of `native' gates, and some gates are less prone to errors than others when implemented. 
Hence, on NISQ machines the choice of ansatz should be informed by the actual hardware to be used, to avoid unnecessary gate-depth after compilation and hence noise from mere re-parametrisation.

\textit{Step 5 (Quantum Compiler):} A quantum compiler translates the quantum circuit into machine specific instructions. 
This includes expressing the circuit's quantum gates in terms of those actually physically available on that machine and `routing' the qubits to allow for the necessary interactions given the device's topology, as well as an optimisation to further reduce noise.

\ed{\textit{Step 6 (Quantum Computer):} The quantum computer runs the circuit.} 
\ed{In order to see what this means more precisely, recall that the 0-effects in Fig.~\ref{fig:ExampleSent_QCircuit}, while crucial parts of the sentence's representation, are not deterministically implementable operations; as outcomes of measurements, they are obtained only with a certain probability. 
The circuit, which corresponds to that of Fig.~\ref{fig:ExampleSent_QCircuit}, but that can actually be implemented, is hence precisely that of Fig.~\ref{fig:Example_circuit} with the additional operation of measuring all 5 qubits at the end.} 

\ed{Thus the quantum computer runs a given circuit $n_{\text{shots}}$ times (runs also referred to as \textit{shots}), but ignoring the 0-effects that appear in a sentence's representation for the time being. 
For each run the machine prepares initial states, applies the gates and then measures all the qubits. At the end, this step returns outcome counts of the shots for \emph{all} qubits.}

\ed{\textit{Step 7 (Post-Processing):} 
The post-processing step of the pipeline has two parts to it. 
The first is a task independent one and is the completion of what it means to implement a sentence's circuit representation like the one in Fig.~\ref{fig:ExampleSent_QCircuit}, that is, taking into account where there are 0-effects (or more generally other non-deterministic effects).}

\ed{This thus is the point at which \emph{post-selection} happens, a concept that was introduced in Sec.~\ref{sec:Intro_QC}.   
As part of \emph{Step 6} all qubits were measured yielding counts of outcomes. 
Post-selection now means to restrict the measurement data to the subset of runs where all those qubits with a 0-effect on them in the circuit to be achieved, did indeed yield the corresponding 0 outcome when measured. 
Only once the data has been appropriately restricted in such manner are the counts turned into estimations of relative frequencies for the outcomes of those qubits that correspond to the `open' wires in the run circuit, such as the wire coming out of component (i) in Fig.~\ref{fig:ExampleSent_QCircuit}.} 

\ed{So the business of post-selection is a simple matter of conditioning! However, we now finally see the reason for why cups are costly: each cup leads to $2q_n$ (or $2q_s$ ) qubits that require post-selection. Had one stuck to the diagram in Fig.~\ref{fig:ExampleSent_PGDiagram} then 6 out of 7 qubits would have had to be post-selected, rather than 4 out of 5 based on Fig.~\ref{fig:ExampleSent_PGDiagram_bent}. 
If all possible outcomes were uniformly distributed, even just post-selecting 4 qubits to yield the outcome 0, would mean that only one in $2^4$ shots meets the condition. 
Depending on the actual values of the parameters for which a parametrised circuit is run, if post-selecting on $p$ qubits the probability for seeing the $p$ qubits to yield outcome 0 can be significantly lower (or larger of course) than $1/(2^p)$ and will vary as a training algorithm varies the parameters. 
Hence, with sentences slightly longer than in our running example and limited number of shots\footnote{\ed{For IBM quantum devices the maximum number of shots was $2^{13}$ at the time of running this experiment. Then post-selecting 6 out of 7 (4 out of 5) qubits and assuming a uniform distribution means that $2^7$ ($2^9$) runs give counts over the remaining qubit whose outcome distribution one wants to estimate. 
Just to reiterate: for any specific choice of parameters in the circuit the distribution is likely not to be uniform and the runs that meet the conditions can be many fewer or more. Generically, nothing much can be said; we only note that in our experimental set-up explained in detail in Sec.~\protect\ref{sec:Experiments} post-selecting 6 qubits sometimes would leave only very few (of the order of 10) useable shots, whereas post-selecting only 4 always yielded sufficient statistics -- hence, the presented rewrite procedure does the job for the purpose of this work.}} 
of any given circuit one easily runs into severe statistical limitations to reliably estimate the desired outcome probabilities. 
The method of `bending down nouns' (\emph{Step 4}) therefore is an effective and useful rewrite procedure for the models this work studies.}

\ed{Lastly, the second part of this \emph{Step 7} takes the relative frequencies as input and completes any further post-processing on them for the calculation of a \emph{task-specific} final result such as the computation of a cost function.}

\subsection{The Pipeline Adjusted to the Word-Sequence Model}
\label{sec:pipeline_ws_model}

The pipeline described above also applies to the \ws~model with the following simplifications. 

\emph{Step 1} is irrelevant since no parse tree is input into this model. 

In \emph{Step 2}, instead of a \discocat~diagram based on a parse-tree, a string diagram of the form shown in Figure~\ref{fig:WordSeqModel_as_StringDiagram} is created. Following a state with a single output wire for the `start' token on the very left, every word is represented by a state with two output wires and then every adjacent pair of states is connected up with a cup. 
For the example sentence of this section the corresponding diagram is shown in Figure~\ref{fig:ExampleSentenceWordSeqModel}. 
Note that the wires have no explicit typing that would reflect the words' composite grammatical types like in the syntax-based \discocat~model. 
However, in light of the intended interpretation of the diagram in the category of (complex) vector spaces, all wires are to be treated as of the same kind. For the remainder of the pipeline to go through the wires thus have to be typed consistently with, say, $n$ -- it does not matter what the label is as long as it is the same for all of them.   

Concerning \emph{Step 3}, since this work will only present classical simulations for this model rather than an actual quantum implementation, there is no practical need to fix any particular rewrite procedure to optimise the string diagrams. We only note that this could easily be done.\footnote{A suitable choice would e.g. be to start from the right of the diagram and flip every other state into an effect according to the \emph{bigraph} recipe from \shortcite{MeichanetzidisEtAl_2020_QNLPOnNISQ}.} 

Finally, concerning \emph{Steps 4-7}, these apply in the same way as for \discocat. 
In particular, the same choices constituting an ansatz for the \discocat~model in \emph{Step 4} specify a map that turns any \ws~string diagram into a specific quantum circuit. (NB the above comment on treating all wires as labelled consistently with, say, $n$).

\vspace*{0.3cm}
\begin{figure*}[!h]
	\centering			
	\begin{subfigure}{0.47\textwidth}
		\centering
		
\begin{tikzpicture}
	\begin{pgfonlayer}{nodelayer}
		\node [style=none] (118) at (-11.25, -7) {};
		\node [style=none] (119) at (-12, -8.75) {};
		\node [style=none] (120) at (-7.5, -8.75) {};
		\node [style=none] (121) at (-8.25, -7) {};
		\node [style=none] (122) at (-9.75, -8) {\small{person}};
		\node [style=none] (123) at (-6, -7) {};
		\node [style=none] (124) at (-6.75, -8.75) {};
		\node [style=none] (125) at (-1.25, -8.75) {};
		\node [style=none] (126) at (-2, -7) {};
		\node [style=none] (127) at (-4, -8) {\small{prepares}};
		\node [style=none] (128) at (0.25, -7) {};
		\node [style=none] (129) at (-0.5, -8.75) {};
		\node [style=none] (130) at (4, -8.75) {};
		\node [style=none] (131) at (3.25, -7) {};
		\node [style=none] (132) at (1.75, -8) {\small{tasty}};
		\node [style=none] (133) at (5.5, -7) {};
		\node [style=none] (134) at (4.75, -8.75) {};
		\node [style=none] (135) at (9.25, -8.75) {};
		\node [style=none] (136) at (8.5, -7) {};
		\node [style=none] (137) at (7, -8) {\small{dinner}};
		\node [style=none] (138) at (-15.25, -7) {};
		\node [style=none] (139) at (-16, -8.75) {};
		\node [style=none] (140) at (-12.75, -8.75) {};
		\node [style=none] (141) at (-13.5, -7) {};
		\node [style=none] (142) at (-14.5, -8) {\small{$\langle S \rangle$}};
		\node [style=none] (143) at (2.5, -8.75) {};
		\node [style=none] (144) at (6.5, -8.75) {};
		\node [style=none] (145) at (-3, -8.75) {};
		\node [style=none] (146) at (1, -8.75) {};
		\node [style=none] (147) at (-9, -8.75) {};
		\node [style=none] (148) at (-5, -8.75) {};
		\node [style=none] (149) at (-14.5, -8.75) {};
		\node [style=none] (150) at (-10.5, -8.75) {};
		\node [style=none] (151) at (7.75, -8.75) {};
		\node [style=none] (152) at (7.75, -11) {};
	\end{pgfonlayer}
	\begin{pgfonlayer}{edgelayer}
		\draw (119.center) to (118.center);
		\draw (118.center) to (121.center);
		\draw (121.center) to (120.center);
		\draw (119.center) to (120.center);
		\draw (124.center) to (123.center);
		\draw (123.center) to (126.center);
		\draw (126.center) to (125.center);
		\draw (124.center) to (125.center);
		\draw (129.center) to (128.center);
		\draw (128.center) to (131.center);
		\draw (131.center) to (130.center);
		\draw (129.center) to (130.center);
		\draw (134.center) to (133.center);
		\draw (133.center) to (136.center);
		\draw (136.center) to (135.center);
		\draw (134.center) to (135.center);
		\draw (139.center) to (138.center);
		\draw (138.center) to (141.center);
		\draw (141.center) to (140.center);
		\draw (139.center) to (140.center);
		\draw [bend right=90, looseness=1.25] (143.center) to (144.center);
		\draw [bend right=90, looseness=1.25] (145.center) to (146.center);
		\draw [bend right=90, looseness=1.25] (147.center) to (148.center);
		\draw [bend right=90, looseness=1.25] (149.center) to (150.center);
		\draw (151.center) to (152.center);
	\end{pgfonlayer}
\end{tikzpicture}}

		\caption{}
		\label{fig:ExampleSentenceWordSeqModel}
	\end{subfigure}
	\hspace*{0.5cm}
	\begin{subfigure}{0.47\textwidth}
		\centering
		
\begin{tikzpicture}
	\begin{pgfonlayer}{nodelayer}
		\node [style=none] (105) at (-0.75, -12.25) {};
		\node [style=none] (106) at (-0.75, -14.25) {};
		\node [style={white_dot, scale=1.6}] (107) at (-0.75, -11.75) {};
		\node [style=none] (110) at (1.75, -8.75) {};
		\node [style=none] (111) at (-0.5, -11.25) {};
		\node [style=none] (112) at (7, -8.75) {};
		\node [style=none] (113) at (-0.25, -11.5) {};
		\node [style=none] (114) at (-4, -8.75) {};
		\node [style=none] (115) at (-1, -11.25) {};
		\node [style=none] (116) at (-9.75, -8.75) {};
		\node [style=none] (117) at (-1.25, -11.5) {};
		\node [style=none] (118) at (-11.25, -7) {};
		\node [style=none] (119) at (-12, -8.75) {};
		\node [style=none] (120) at (-7.5, -8.75) {};
		\node [style=none] (121) at (-8.25, -7) {};
		\node [style=none] (122) at (-9.75, -8) {\small{person}};
		\node [style=none] (123) at (-6, -7) {};
		\node [style=none] (124) at (-6.75, -8.75) {};
		\node [style=none] (125) at (-1.25, -8.75) {};
		\node [style=none] (126) at (-2, -7) {};
		\node [style=none] (127) at (-4, -8) {\small{prepares}};
		\node [style=none] (128) at (0.25, -7) {};
		\node [style=none] (129) at (-0.5, -8.75) {};
		\node [style=none] (130) at (4, -8.75) {};
		\node [style=none] (131) at (3.25, -7) {};
		\node [style=none] (132) at (1.75, -8) {\small{tasty}};
		\node [style=none] (133) at (5.5, -7) {};
		\node [style=none] (134) at (4.75, -8.75) {};
		\node [style=none] (135) at (9.25, -8.75) {};
		\node [style=none] (136) at (8.5, -7) {};
		\node [style=none] (137) at (7, -8) {\small{dinner}};
		\node [style=none] (138) at (-9.75, -8.75) {};
		\node [style=none] (140) at (-4, -8.75) {};
		\node [style=none] (143) at (7, -8.75) {};
	\end{pgfonlayer}
	\begin{pgfonlayer}{edgelayer}
		\draw (105.center) to (106.center);
		\draw [in=60, out=-90, looseness=0.75] (110.center) to (111.center);
		\draw [in=30, out=-90, looseness=0.75] (112.center) to (113.center);
		\draw [in=120, out=-90, looseness=0.75] (114.center) to (115.center);
		\draw [in=135, out=-90, looseness=0.45] (116.center) to (117.center);
		\draw (119.center) to (118.center);
		\draw (118.center) to (121.center);
		\draw (121.center) to (120.center);
		\draw (119.center) to (120.center);
		\draw (124.center) to (123.center);
		\draw (123.center) to (126.center);
		\draw (126.center) to (125.center);
		\draw (124.center) to (125.center);
		\draw (129.center) to (128.center);
		\draw (128.center) to (131.center);
		\draw (131.center) to (130.center);
		\draw (129.center) to (130.center);
		\draw (134.center) to (133.center);
		\draw (133.center) to (136.center);
		\draw (136.center) to (135.center);
		\draw (134.center) to (135.center);
	\end{pgfonlayer}
\end{tikzpicture}}

		\caption{}
		\label{fig:ExampleSentenceBagofWordsModel}
	\end{subfigure}
	\caption{The string diagram representation of step 2 for the example sentence according to the \ws~model in (a) and according to the \bow~model in (b), respectively.}
	\label{fig:ExampleSentenceBaselineModles}
\end{figure*}

\rl{While the pipeline thus is general and can instantiate the \ws~model, too, we note that for this baseline model we will present experimental results from classical simulations (see Sec.~\ref{subsec:Simulation}), but not experiments from an implementation on a quantum computer. See Sec.~\ref{subsec:QuantumRuns} for the reasoning behind this choice.}

\subsection{The Pipeline Adjusted to the Bag-of-Words Model}
\label{sec:pipeline_bow_model}

The same simplifications of the pipeline laid out for the \ws~model, also apply for the \bow~model. The only difference is that in \emph{Step 2} the construction of the string diagram for a sentence follows the recipe for this model: every word is represented by a state with a single output wire which are all connected with a single `merge-dot'. 
Figure~\ref{fig:ExampleSentenceBagofWordsModel} depicts the resulting diagram in the \bow~model for the example sentence.  
Note that when specifying an ansatz in \emph{Step 4}, just as a cup has a fixed meaning in the resultant quantum circuit \ed{(see the discussion of \emph{Step 4} in the general pipeline above)}, so does the merge-dot have a fixed meaning (given a fixed basis of the underlying vector spaces). 
\ed{See Sec.~\ref{sec:basline_models} for the definition of the map represented by the merge-dot and see Section~\ref{sec:Experiments} for details on how to realise this map explicitly through a choice of quantum gates.} 

\rl{Again, this establishes the generality of the pipeline, allowing us to treat all three kinds of models in a unified way, but just as for the \ws~model, also for this baseline model we will present only experimental results from classical simulations.}

\section{The Tasks}
\label{sec:TheTasks}

We define two simple binary classification tasks for sentences. In the first one, we generated sentences of simple syntactical forms (containing 3-4 words) from a fixed vocabulary by using a simple context-free grammar (Table \ref{tbl:mc_grammar}). 
\begin{table}[h]
\small
\center
\begin{tabular}{l}
\hline
  \texttt{noun\_phrase} $\to$ \texttt{noun} \\
  \texttt{noun\_phrase} $\to$ \texttt{adjective} \texttt{noun} \\
  \texttt{verb\_phrase} $\to$ \texttt{verb} \texttt{noun\_phrase} \\
  \texttt{sentence} $\to$ \texttt{noun\_phrase} \texttt{verb\_phrase} \\
\hline
\end{tabular}
\caption{Context-free grammar for the \textit{MC} task.}
\label{tbl:mc_grammar}
\end{table} 
The nature of the vocabulary (of size 17) allowed us to choose sentences that look natural and refer to one of two possible topics, food or IT. The chosen dataset of this task, henceforth referred to as \MC\ (`meaning classification'), consists of 65 sentences from each topic, similar to the following:
\begin{quote}
  ``skillful programmer creates software'' \\
  ``chef prepares delicious meal''
\end{quote}
Part of the vocabulary \ed{(four words)} is shared between the two classes, so the task (while still an easy one from an NLP perspective) is not trivial. 

In a slightly more conceptual task, we select 105 noun phrases containing relative clauses from the \textsc{RelPron} dataset \cite{relpron}. The phrases are selected in such a way that each word occurs at least 3 times in the dataset, yielding an overall vocabulary of 115 words. While the original task is to map textual definitions (such as ``device that detects planets'') to terms (``telescope''), for the purposes of this work we convert the problem into a binary-classification one, with the goal to predict whether a certain noun phrase contains a subject relative clause (``device that detects planets'') or an object relative clause (``device that observatory has''). Our motivation behind this task, henceforth referred to as \RP ('relative pronoun'), is that it requires some syntax sensitivity from the model, so it is a reasonable choice for testing the \discocat\ model. In addition, the size of vocabulary and consequently the sparseness of words make this task a much more challenging benchmark compared to the \MC~task. 

These simple datasets already pose challenges in two ways.  
First, concerning the lengths of sentences (see \textit{Step 5} of Sec.~\ref{sec:pipeline} and also Sec.~\ref{sec:Conclusion}). 
Second, concerning the lengths of datasets, since they already reach the limits of the currently available quantum hardware -- even just doubling the number of sentences would start to approach an unfeasible time cost given the shared available resources (more details about this in Sec.~\ref{subsec:QuantumRuns}).

\section{Experiments}
\label{sec:Experiments}

The experiments reported in this paper address the tasks \MC\ and \RP\ by implementing the pipeline from Fig.~\ref{fig:pipeline} together with an optimisation procedure to train the model parameters against an objective function, as in standard supervised machine learning. 
Importantly, for all models that this work discusses, training the model amounts to learning the word representations in a task-specific way. 
Recall that the specificities of these word representations -- the size of the embedding space and how the parameters define the word representations -- are contingent on the choice of hyperparameters that fix the ansatz, while all other aspects of the sentence or phrase representations are dictated by syntax (to the degree that the model accounts for syntax at all). 

As explained in Section~\ref{sec:classical_model_overview} the \discocat~model is the most involved, expressive and interesting model for the purpose of this paper. 
The \bow\ and the \ws~models take the role of baseline models in order to show through a comparison that, given the different natures of the  \MC\ and \RP~tasks, the three distinct models' performances and behaviours are as expected and well understood. All models are simulated and in addition the \discocat\ model is implemented on real quantum hardware. 

Our Python implementation\footnote{The Python code and the datasets are available at \url{https://github.com/CQCL/qnlp_lorenz_etal_2021_resources}.} used \texttt{lambeq}\footnote{\url{https://cqcl.github.io/lambeq} \label{footnote:github_repo}} and the \textsc{DisCoPy}  package\footnote{\url{https://github.com/discopy}} \shortcite{discopy} to implement the model specific \textit{Steps 2-4}, the Python interface of the quantum compiler TKET$^{\text{TM}}$ \footnote{\url{https://github.com/CQCL/pytket}} \shortcite{SivarajahEtAl_2020_TKetRetargetableCompiler} for \textit{Step 5}, and the IBM device \texttt{ibmq\_bogota} for \textit{Step 6}. The remainder of this section describes all steps in detail. 

For the first step of parsing sentences, necessary for the \discocat~model, Section~\ref{sec:pipeline} noted that due to the simplicity of both tasks' datasets this parsing can actually be done semi-automatically. For the noun phrases and sentences considered in this work note that relative pronouns in the subject case take pregroup type $n^r \cdot n \cdot s^l \cdot n$, while in the object case their type is $n^r \cdot n \cdot (n^l)^l \cdot s^l$, and that the types for adjectives and transitive verbs are $n \cdot n^l$ and $n^r \cdot s \cdot n^l$, respectively. 
\ed{These are the only pregroup types needed across both our datasets, in order to work out the derivations (see Sec.~\ref{sec:discocat} and \textit{Step 2} in Sec.~\ref{sec:pipeline}) -- one can easily convince oneself that all phrases and sentences in our datasets then have a unique reduction to $n$ or $s$, respectively. 
For an example of how to implement the parsing step in a semi-automatic way simply using a look-up table with the very few different sentence structures appearing in the \MC\ and \RP\ datasets, see the github repository in footnote~\ref{footnote:github_repo}. 
In order to illustrate this point, note that the \discocat\ diagrams in Figures~\ref{fig:ExampleFromRP_subj} and \ref{fig:ExampleFromRP_obj} depict two noun phrases from the \RP\ dataset. 
The corresponding representations according to the baseline models for the first of these two examples can be found in Figures~\ref{fig:ExampleSentenceWordSeqModel_2} and \ref{fig:ExampleSentenceBagofWordsModel_2}. 
Furthermore, the example sentence that was used in Section~\ref{sec:pipeline} is in fact from the \MC\ dataset -- Figure~\ref{fig:ExampleSent_PGDiagram} showed it for the \discocat~model and Figure~\ref{fig:ExampleSentenceBaselineModles} for the baseline models.}  
The scheme of `bending the nouns', described in \textit{Step 3} of Section~\ref{sec:pipeline}, was consistently applied to both datasets in case of the  \discocat~model, while skipped for the baseline models (see Section~\ref{sec:pipeline} for an explanation). 

\begin{figure}[h]
	\centering
	\begin{minipage}{0.48\textwidth}
		\begin{subfigure}{0.95\textwidth}
			\centering
			
\begin{tikzpicture}
	\begin{pgfonlayer}{nodelayer}
		\node [style=none] (118) at (-11.5, -7) {};
		\node [style=none] (119) at (-12.25, -8.75) {};
		\node [style=none] (120) at (-7.75, -8.75) {};
		\node [style=none] (121) at (-8.5, -7) {};
		\node [style=none] (122) at (-10, -8) {\small{device}};
		\node [style=none] (123) at (-5.5, -7) {};
		\node [style=none] (124) at (-6.25, -8.75) {};
		\node [style=none] (125) at (-1.5, -8.75) {};
		\node [style=none] (126) at (-2.25, -7) {};
		\node [style=none] (127) at (-3.75, -8) {\small{that}};
		\node [style=none] (128) at (1, -7) {};
		\node [style=none] (129) at (0.25, -8.75) {};
		\node [style=none] (130) at (4.75, -8.75) {};
		\node [style=none] (131) at (4, -7) {};
		\node [style=none] (132) at (2.5, -8) {\small{detects}};
		\node [style=none] (133) at (6.75, -7) {};
		\node [style=none] (134) at (6, -8.75) {};
		\node [style=none] (135) at (10.5, -8.75) {};
		\node [style=none] (136) at (9.75, -7) {};
		\node [style=none] (137) at (8.25, -8) {\small{planet}};
		\node [style=none] (138) at (3.75, -8.75) {};
		\node [style=none] (139) at (8.25, -8.75) {};
		\node [style=none] (140) at (4.75, -9.5) {\tiny{$n^l$}};
		\node [style=none] (141) at (8.75, -9.5) {\tiny{$n$}};
		\node [style=none] (142) at (-10.25, -8.75) {};
		\node [style=none] (143) at (-5.5, -8.75) {};
		\node [style=none] (144) at (-10.75, -9.5) {\tiny{$n$}};
		\node [style=none] (145) at (-6.5, -9.5) {\tiny{$n^r$}};
		\node [style=none] (146) at (-2.5, -8.75) {};
		\node [style=none] (147) at (1.5, -8.75) {};
		\node [style=none] (148) at (-1.5, -9.5) {\tiny{$n$}};
		\node [style=none] (149) at (0.5, -9.5) {\tiny{$n^r$}};
		\node [style=none] (150) at (-4.75, -8.75) {};
		\node [style=none] (151) at (-4.75, -12) {};
		\node [style=none] (152) at (-5.25, -11.25) {\tiny{$n$}};
		\node [style=none] (153) at (-3.5, -8.75) {};
		\node [style=none] (154) at (2.5, -8.75) {};
		\node [style=none] (155) at (-3.75, -10.5) {\tiny{$s^l$}};
		\node [style=none] (156) at (2.75, -9.75) {\tiny{$s$}};
	\end{pgfonlayer}
	\begin{pgfonlayer}{edgelayer}
		\draw (119.center) to (118.center);
		\draw (118.center) to (121.center);
		\draw (121.center) to (120.center);
		\draw (119.center) to (120.center);
		\draw (124.center) to (123.center);
		\draw (123.center) to (126.center);
		\draw (126.center) to (125.center);
		\draw (124.center) to (125.center);
		\draw (129.center) to (128.center);
		\draw (128.center) to (131.center);
		\draw (131.center) to (130.center);
		\draw (129.center) to (130.center);
		\draw (134.center) to (133.center);
		\draw (133.center) to (136.center);
		\draw (136.center) to (135.center);
		\draw (134.center) to (135.center);
		\draw [bend right=90, looseness=1.50] (138.center) to (139.center);
		\draw [bend right=90, looseness=1.50] (142.center) to (143.center);
		\draw [bend right=90, looseness=1.50] (146.center) to (147.center);
		\draw (150.center) to (151.center);
		\draw [bend right=90, looseness=1.50] (153.center) to (154.center);
	\end{pgfonlayer}
\end{tikzpicture}}

			\vspace*{-0.3cm}
			\caption{}
			\label{fig:ExampleFromRP_subj}
		\end{subfigure}
		\begin{subfigure}{0.95\textwidth}
			\centering
			
\begin{tikzpicture}
	\begin{pgfonlayer}{nodelayer}
		\node [style=none] (118) at (-10.75, -7) {};
		\node [style=none] (119) at (-11.5, -8.75) {};
		\node [style=none] (120) at (-7, -8.75) {};
		\node [style=none] (121) at (-7.75, -7) {};
		\node [style=none] (122) at (-9.25, -8) {\small{device}};
		\node [style=none] (123) at (-5.5, -7) {};
		\node [style=none] (124) at (-6.25, -8.75) {};
		\node [style=none] (125) at (-1.5, -8.75) {};
		\node [style=none] (126) at (-2.25, -7) {};
		\node [style=none] (127) at (-3.75, -8) {\small{that}};
		\node [style=none] (128) at (0, -7) {};
		\node [style=none] (129) at (-0.75, -8.75) {};
		\node [style=none] (130) at (6.25, -8.75) {};
		\node [style=none] (131) at (5.5, -7) {};
		\node [style=none] (132) at (2.75, -8) {\small{observatory}};
		\node [style=none] (133) at (7.75, -7) {};
		\node [style=none] (134) at (7, -8.75) {};
		\node [style=none] (135) at (10.75, -8.75) {};
		\node [style=none] (136) at (10, -7) {};
		\node [style=none] (137) at (9, -8) {\small{has}};
		\node [style=none] (138) at (2.75, -8.75) {};
		\node [style=none] (139) at (7.75, -8.75) {};
		\node [style=none] (140) at (10.5, -9.5) {\tiny{$n^l$}};
		\node [style=none] (141) at (-3.25, -11.5) {\tiny{$(n^l)^l$}};
		\node [style=none] (142) at (-9.5, -8.75) {};
		\node [style=none] (143) at (-5.5, -8.75) {};
		\node [style=none] (144) at (-10.25, -9.25) {\tiny{$n$}};
		\node [style=none] (145) at (-6.5, -9.5) {\tiny{$n^r$}};
		\node [style=none] (146) at (-2.25, -8.75) {};
		\node [style=none] (147) at (9, -8.75) {};
		\node [style=none] (148) at (2.25, -9.25) {\tiny{$n$}};
		\node [style=none] (149) at (6.75, -9.5) {\tiny{$n^r$}};
		\node [style=none] (150) at (-4.75, -8.75) {};
		\node [style=none] (151) at (-4.75, -13.5) {};
		\node [style=none] (152) at (-5.25, -12.75) {\tiny{$n$}};
		\node [style=none] (153) at (-3.5, -8.75) {};
		\node [style=none] (154) at (10, -8.75) {};
		\node [style=none] (155) at (-1.25, -9.5) {\tiny{$s^l$}};
		\node [style=none] (156) at (8.25, -9.25) {\tiny{$s$}};
	\end{pgfonlayer}
	\begin{pgfonlayer}{edgelayer}
		\draw (119.center) to (118.center);
		\draw (118.center) to (121.center);
		\draw (121.center) to (120.center);
		\draw (119.center) to (120.center);
		\draw (124.center) to (123.center);
		\draw (123.center) to (126.center);
		\draw (126.center) to (125.center);
		\draw (124.center) to (125.center);
		\draw (129.center) to (128.center);
		\draw (128.center) to (131.center);
		\draw (131.center) to (130.center);
		\draw (129.center) to (130.center);
		\draw (134.center) to (133.center);
		\draw (133.center) to (136.center);
		\draw (136.center) to (135.center);
		\draw (134.center) to (135.center);
		\draw [bend right=90, looseness=1.25] (138.center) to (139.center);
		\draw [bend right=90, looseness=1.50] (142.center) to (143.center);
		\draw [bend right=90, looseness=0.75] (146.center) to (147.center);
		\draw (150.center) to (151.center);
		\draw [bend right=90] (153.center) to (154.center);
	\end{pgfonlayer}
\end{tikzpicture}}

			\caption{}
			\label{fig:ExampleFromRP_obj}
		\end{subfigure}
	\end{minipage}
	\begin{minipage}{0.48\textwidth}
		\begin{subfigure}{0.95\textwidth}
			\centering
			
\begin{tikzpicture}
	\begin{pgfonlayer}{nodelayer}
		\node [style=none] (118) at (-11.25, -7) {};
		\node [style=none] (119) at (-12, -8.75) {};
		\node [style=none] (120) at (-7.5, -8.75) {};
		\node [style=none] (121) at (-8.25, -7) {};
		\node [style=none] (122) at (-9.75, -8) {\small{device}};
		\node [style=none] (123) at (-5.75, -7) {};
		\node [style=none] (124) at (-6.5, -8.75) {};
		\node [style=none] (125) at (-2, -8.75) {};
		\node [style=none] (126) at (-2.75, -7) {};
		\node [style=none] (127) at (-4.25, -8) {\small{that}};
		\node [style=none] (128) at (-0.5, -7) {};
		\node [style=none] (129) at (-1.25, -8.75) {};
		\node [style=none] (130) at (3.25, -8.75) {};
		\node [style=none] (131) at (2.5, -7) {};
		\node [style=none] (132) at (1, -8) {\small{detects}};
		\node [style=none] (133) at (4.75, -7) {};
		\node [style=none] (134) at (4, -8.75) {};
		\node [style=none] (135) at (8.5, -8.75) {};
		\node [style=none] (136) at (7.75, -7) {};
		\node [style=none] (137) at (6.25, -8) {\small{planet}};
		\node [style=none] (138) at (-15.25, -7) {};
		\node [style=none] (139) at (-16, -8.75) {};
		\node [style=none] (140) at (-12.75, -8.75) {};
		\node [style=none] (141) at (-13.5, -7) {};
		\node [style=none] (142) at (-14.5, -8) {\small{$\langle S \rangle$}};
		\node [style=none] (143) at (1.75, -8.75) {};
		\node [style=none] (144) at (5.75, -8.75) {};
		\node [style=none] (145) at (-3.5, -8.75) {};
		\node [style=none] (146) at (0.25, -8.75) {};
		\node [style=none] (147) at (-9, -8.75) {};
		\node [style=none] (148) at (-5, -8.75) {};
		\node [style=none] (149) at (-14.5, -8.75) {};
		\node [style=none] (150) at (-10.5, -8.75) {};
		\node [style=none] (151) at (7, -8.75) {};
		\node [style=none] (152) at (7, -11) {};
	\end{pgfonlayer}
	\begin{pgfonlayer}{edgelayer}
		\draw (119.center) to (118.center);
		\draw (118.center) to (121.center);
		\draw (121.center) to (120.center);
		\draw (119.center) to (120.center);
		\draw (124.center) to (123.center);
		\draw (123.center) to (126.center);
		\draw (126.center) to (125.center);
		\draw (124.center) to (125.center);
		\draw (129.center) to (128.center);
		\draw (128.center) to (131.center);
		\draw (131.center) to (130.center);
		\draw (129.center) to (130.center);
		\draw (134.center) to (133.center);
		\draw (133.center) to (136.center);
		\draw (136.center) to (135.center);
		\draw (134.center) to (135.center);
		\draw (139.center) to (138.center);
		\draw (138.center) to (141.center);
		\draw (141.center) to (140.center);
		\draw (139.center) to (140.center);
		\draw [bend right=90, looseness=1.25] (143.center) to (144.center);
		\draw [bend right=90, looseness=1.50] (145.center) to (146.center);
		\draw [bend right=90, looseness=1.25] (147.center) to (148.center);
		\draw [bend right=90, looseness=1.25] (149.center) to (150.center);
		\draw (151.center) to (152.center);
	\end{pgfonlayer}
\end{tikzpicture}}

			\vspace*{-0.3cm}
			\caption{}
			\label{fig:ExampleSentenceWordSeqModel_2}
		\end{subfigure}
		\begin{subfigure}{0.95\textwidth}
			\centering	
			
\begin{tikzpicture}
	\begin{pgfonlayer}{nodelayer}
		\node [style=none] (105) at (-0.75, -12.25) {};
		\node [style=none] (106) at (-0.75, -14.25) {};
		\node [style={white_dot, scale=1.6}] (107) at (-0.75, -11.75) {};
		\node [style=none] (110) at (1.75, -8.75) {};
		\node [style=none] (111) at (-0.5, -11.25) {};
		\node [style=none] (112) at (7, -8.75) {};
		\node [style=none] (113) at (-0.25, -11.5) {};
		\node [style=none] (114) at (-3.75, -8.75) {};
		\node [style=none] (115) at (-1, -11.25) {};
		\node [style=none] (116) at (-9.25, -8.75) {};
		\node [style=none] (117) at (-1.25, -11.5) {};
		\node [style=none] (118) at (-10.75, -7) {};
		\node [style=none] (119) at (-11.5, -8.75) {};
		\node [style=none] (120) at (-7, -8.75) {};
		\node [style=none] (121) at (-7.75, -7) {};
		\node [style=none] (122) at (-9.25, -8) {\small{device}};
		\node [style=none] (123) at (-5.5, -7) {};
		\node [style=none] (124) at (-6.25, -8.75) {};
		\node [style=none] (125) at (-1.25, -8.75) {};
		\node [style=none] (126) at (-2, -7) {};
		\node [style=none] (127) at (-3.75, -8) {\small{that}};
		\node [style=none] (128) at (0.25, -7) {};
		\node [style=none] (129) at (-0.5, -8.75) {};
		\node [style=none] (130) at (4, -8.75) {};
		\node [style=none] (131) at (3.25, -7) {};
		\node [style=none] (132) at (1.75, -8) {\small{detects}};
		\node [style=none] (133) at (5.5, -7) {};
		\node [style=none] (134) at (4.75, -8.75) {};
		\node [style=none] (135) at (9.25, -8.75) {};
		\node [style=none] (136) at (8.5, -7) {};
		\node [style=none] (137) at (7, -8) {\small{planet}};
		\node [style=none] (138) at (-9.25, -8.75) {};
		\node [style=none] (140) at (-3.75, -8.75) {};
		\node [style=none] (143) at (7, -8.75) {};
	\end{pgfonlayer}
	\begin{pgfonlayer}{edgelayer}
		\draw (105.center) to (106.center);
		\draw [in=60, out=-90, looseness=0.75] (110.center) to (111.center);
		\draw [in=30, out=-90, looseness=0.75] (112.center) to (113.center);
		\draw [in=120, out=-90, looseness=0.75] (114.center) to (115.center);
		\draw [in=135, out=-90, looseness=0.45] (116.center) to (117.center);
		\draw (119.center) to (118.center);
		\draw (118.center) to (121.center);
		\draw (121.center) to (120.center);
		\draw (119.center) to (120.center);
		\draw (124.center) to (123.center);
		\draw (123.center) to (126.center);
		\draw (126.center) to (125.center);
		\draw (124.center) to (125.center);
		\draw (129.center) to (128.center);
		\draw (128.center) to (131.center);
		\draw (131.center) to (130.center);
		\draw (129.center) to (130.center);
		\draw (134.center) to (133.center);
		\draw (133.center) to (136.center);
		\draw (136.center) to (135.center);
		\draw (134.center) to (135.center);
	\end{pgfonlayer}
\end{tikzpicture}}

			\caption{}
			\label{fig:ExampleSentenceBagofWordsModel_2}
		\end{subfigure}
	\end{minipage}
	\caption{\discocat~diagrams for example phrases from \rl{the \textsc{RelPron} dataset \cite{relpron}, used in the \RP\ task,}
	where in (a) `device' is the subject of the verb, while in (b) it is the object; 
	(c) and (d) show the corresponding diagrams for the sentence from (a), but in the \ws\ and the \bow\ models, respectively.}
	\label{fig:MoreExampleDiags}
\end{figure}

As emphasised in Section~\ref{sec:pipeline}, after the first three (model specific) steps the remaining steps of the pipeline are the same for each model.  After \textit{Step 3} the sentences (or noun phrases) for all models are now represented by string diagrams, waiting to be translated into quantum circuits. 
As mentioned previously, it is convenient -- as a mere matter of bookkeeping -- to assign the type $n$ to all wires of a diagram in the \bow\ and the \ws~models independently from whether the diagram represents a sentence or a noun phrase. This allows for the same hyperparameters that determine an ansatz for the \discocat~model to also determine the ansatz for the baseline models in a way that ensures the number of parameters to be comparable.

\subsection{Model Parametrisation and \Ansaetze}
\label{subsec:Parametrisation}

The exposition of \textit{Step 4} in Section~\ref{sec:pipeline} explained how the choice of ansatz determines the parametrisation of a concrete family of models. We studied a variety of \ansaetze\ for both tasks. For each ansatz all appearing parameters are valued in $[0, 2\pi]$. 

In order to keep the number of qubits as low as possible in light of the noise in NISQ devices, while having at least one qubit to encode the label (we are trying to solve a binary classification task after all) we set $q_n=1$ and $q_s = 1$ for the \MC\ task, but $q_s = 0$ for the \RP\ task (noting that the type of the phrases here is $n$ for all three models). 
Recall that the vector spaces in which the noun and sentence embeddings live have (complex) dimension $N=2^{q_n}$ and $S=2^{q_s}$, respectively.
Once the dimensions of the vector spaces that the wires represent are thus fixed, a principled way has to be provided for how to assign concrete (parametrised) quantum states and effects to all appearing word boxes. 

For single qubit states $\ket{w}$ (and their corresponding effects $\bra{w}$) two options were considered. 
\ed{First, an \emph{Euler parametrisation}, $\ket{w} = Rx(\theta_3) Rz(\theta_2) \allowbreak Rx(\theta_1) \ket{0}$. 
The name is owed to the well-known Euler parametrisation of a general rotation in 3-dimensional space ($\mathbb{R}^3$), but all that matters here is that it is a choice of a fully general parametrisation of any single qubit state and that, as a sequence of three rotation gates, it hence involves three parameters that are angles, $\theta_1, \theta_2, \theta_3 \in [0, 2\pi]$. 
See Fig.~\ref{fig:Euler} for the corresponding diagrammatic representation of such an Euler parametrisation of a single qubit state, as well as Sec.~\ref{sec:Intro_QC} and in particular Fig.~\ref{Fig_Def_Basic_gates} for the details on the involved gates.}
Second, the use of a single $Rx$ gate by assigning $\ket{w} = Rx(\theta) \ket{0}$, which, with a single parameter, 
gives a more economical option that is still well-motivated, since an $Rx$ gate can mediate between the $\ket{0}$ and $\ket{1}$ basis states. 
Let $p_n\in\{3,1\}$ represent the choice of which of these two options is chosen. 
Any chosen of the two options is then always consistently applied to all words of the dataset with a single qubit representation. Note that for the \discocat\ model, due to our scheme from \textit{Step 3} (see Sec.~\ref{sec:pipeline}), all nouns appear as single qubit effects $\bra{w}$ and for the \bow~model all words are single qubit states, while for the \ws~model only the start token $\langle S \rangle$ is assigned a single qubit state. 

In the \discocat\ model adjectives are states on two qubits and verbs (only transitive ones appear) are, depending on $q_s$, states on two or three qubits, whereas in the \ws~model all words are two qubit states. 
For such multi-qubit states so called \textit{IQP}\footnote{Instantaneous Quantum Polynomial.}-based states were used. 
\ed{For $m$ qubits such a state consists of all $m$ qubits initialised in $\ket{0}$, followed by $d$ many IQP layers \shortcite{HavlivcekEtAl_2019_SupervisedLearning}. 
Each such layer in turn consists of an $H$ gate on every qubit, subsequently composed with $m-1$ controlled $Rz$ gates, connecting adjacent qubits. For examples see components (i) and (ii) of Fig.~\ref{fig:ExampleSent_QCircuit}, which show one IQP layer for a three and a two qubit state, respectively. 
We stress that the particular choice for the multi qubit state part of an ansatz does not deserve too much significance. One could have chosen some different state parametrisation, i.e. some different well-defined recipe of parametrised layers of gates so long as it satisfies some basic desiderata. The latter are that the qubits can actually interact (at least for most values of the parameters) as is the case in IQP layers through the controlled $Rz$ gates, and that as the number $d$ of layers increases one can (asymptotically) reach any state on the $m$ qubits. 
The particular choice of IQP layers then is, apart from that it satisfies these features, motivated by that it has been studied in the literature, was found to be expressible enough to perform QML \cite{HavlivcekEtAl_2019_SupervisedLearning}, is such that the appearing gates were native to IBMQ's machines at the time the experiments were run and finally, simply because it worked.} 
We considered $d\in \{1,2\}$, again in order to keep the depth of circuits as small as possible. 

\begin{figure}[h]
	\centering
		\begin{subfigure}{0.15\textwidth}
		\centering
		
\begin{tikzpicture}
	\begin{pgfonlayer}{nodelayer}
		\node [style=none] (0) at (-1, 3.25) {};
		\node [style=none] (1) at (0, 4.5) {};
		\node [style=none] (2) at (1, 3.25) {};
		\node [style=none] (6) at (0, 3.25) {};
		\node [style=none] (23) at (0, 3.75) {\tiny{$0$}};
		\node [style=none] (32) at (-1.5, 0.25) {};
		\node [style=none] (33) at (1.5, 0.25) {};
		\node [style=none] (34) at (-1.5, -1.25) {};
		\node [style=none] (35) at (1.5, -1.25) {};
		\node [style=none] (37) at (0, -0.5) {\tiny{$R_z(\theta_2)$}};
		\node [style=none] (38) at (0, 0.25) {};
		\node [style=none] (39) at (0, -1.25) {};
		\node [style=none] (40) at (-1.5, -2) {};
		\node [style=none] (41) at (1.5, -2) {};
		\node [style=none] (42) at (-1.5, -3.5) {};
		\node [style=none] (43) at (1.5, -3.5) {};
		\node [style=none] (45) at (0, -2.75) {\tiny{$R_x(\theta_3)$}};
		\node [style=none] (46) at (0, -2) {};
		\node [style=none] (47) at (0, -3.5) {};
		\node [style=none] (48) at (0, -4.25) {};
		\node [style=none] (78) at (1.5, -0.5) {};
		\node [style=none] (79) at (-1.5, 2.5) {};
		\node [style=none] (80) at (1.5, 2.5) {};
		\node [style=none] (81) at (-1.5, 1) {};
		\node [style=none] (82) at (1.5, 1) {};
		\node [style=none] (83) at (0, 1.75) {\tiny{$R_x(\theta_1)$}};
		\node [style=none] (84) at (0, 2.5) {};
		\node [style=none] (85) at (0, 1) {};
		\node [style=none] (88) at (0, 1) {};
		\node [style=none] (89) at (0, 0.25) {};
		\node [style=none] (90) at (0, 3.25) {};
		\node [style=none] (91) at (0, 2.5) {};
	\end{pgfonlayer}
	\begin{pgfonlayer}{edgelayer}
		\draw (0.center) to (1.center);
		\draw (1.center) to (2.center);
		\draw (0.center) to (2.center);
		\draw (32.center) to (33.center);
		\draw (32.center) to (34.center);
		\draw (34.center) to (35.center);
		\draw (33.center) to (35.center);
		\draw (40.center) to (41.center);
		\draw (40.center) to (42.center);
		\draw (42.center) to (43.center);
		\draw (41.center) to (43.center);
		\draw (39.center) to (46.center);
		\draw (47.center) to (48.center);
		\draw (79.center) to (80.center);
		\draw (79.center) to (81.center);
		\draw (81.center) to (82.center);
		\draw (80.center) to (82.center);
		\draw (88.center) to (89.center);
		\draw (90.center) to (91.center);
	\end{pgfonlayer}
\end{tikzpicture}}

		\caption{}
		\label{fig:Euler}
	\end{subfigure}
	\begin{subfigure}{0.25\textwidth}
		\centering
		
\begin{tikzpicture}
	\begin{pgfonlayer}{nodelayer}
		\node [style=none] (80) at (-3.5, 3.25) {};
		\node [style=none] (81) at (-2.5, 4.5) {};
		\node [style=none] (82) at (-1.5, 3.25) {};
		\node [style=none] (83) at (-2.5, 3.25) {};
		\node [style=none] (84) at (-3, 2.5) {};
		\node [style=none] (85) at (-2, 2.5) {};
		\node [style=none] (86) at (-3, 1.5) {};
		\node [style=none] (87) at (-2, 1.5) {};
		\node [style=none] (88) at (-2.5, 3.75) {\tiny{$0$}};
		\node [style=none] (89) at (-2.5, 2.5) {};
		\node [style=none] (90) at (-2.5, 2) {\tiny{$H$}};
		\node [style=none] (98) at (-2.5, 1.5) {};
		\node [style=none] (101) at (-2.5, -1.75) {};
		\node [style=none] (103) at (-0.75, 3.25) {};
		\node [style=none] (104) at (0.25, 4.5) {};
		\node [style=none] (105) at (1.25, 3.25) {};
		\node [style=none] (106) at (0.25, 3.25) {};
		\node [style=none] (107) at (-0.25, 2.5) {};
		\node [style=none] (108) at (0.75, 2.5) {};
		\node [style=none] (109) at (-0.25, 1.5) {};
		\node [style=none] (110) at (0.75, 1.5) {};
		\node [style=none] (111) at (0.25, 3.75) {\tiny{$0$}};
		\node [style=none] (112) at (0.25, 2.5) {};
		\node [style=none] (113) at (0.25, 2) {\tiny{$H$}};
		\node [style=none] (121) at (0.25, 1.5) {};
		\node [style=none] (127) at (2, 3.25) {};
		\node [style=none] (128) at (3, 4.5) {};
		\node [style=none] (129) at (4, 3.25) {};
		\node [style=none] (130) at (3, 3.25) {};
		\node [style=none] (131) at (2.5, 2.5) {};
		\node [style=none] (132) at (3.5, 2.5) {};
		\node [style=none] (133) at (2.5, 1.5) {};
		\node [style=none] (134) at (3.5, 1.5) {};
		\node [style=none] (135) at (3, 3.75) {\tiny{$0$}};
		\node [style=none] (136) at (3, 2.5) {};
		\node [style=none] (137) at (3, 2) {\tiny{$H$}};
		\node [style=none] (145) at (3, 1.5) {};
		\node [style=none] (160) at (-3, -1.75) {};
		\node [style=none] (161) at (-2, -1.75) {};
		\node [style=none] (162) at (-3, -2.75) {};
		\node [style=none] (163) at (-2, -2.75) {};
		\node [style=none] (165) at (-2.5, -2.25) {\tiny{$H$}};
		\node [style=none] (166) at (-2.5, -2.75) {};
		\node [style=none] (167) at (-2.5, -3.25) {};
		\node [style={white_dot, scale=1.41}] (168) at (-2.5, 0.5) {};
		\node [style={black_dot, scale=1.0}] (169) at (0.25, 0.5) {};
		\node [style=none] (170) at (-3, 0.5) {};
		\node [style=none] (171) at (0.25, 0.5) {};
		\node [style={white_dot, scale=1.41}] (172) at (0.25, -1) {};
		\node [style={black_dot, scale=1.0}] (173) at (3, -1) {};
		\node [style=none] (174) at (-0.25, -1) {};
		\node [style=none] (175) at (3, -1) {};
		\node [style=none] (176) at (0.25, 1.5) {};
		\node [style=none] (177) at (0.25, -3.25) {};
		\node [style=none] (178) at (3, 1.5) {};
		\node [style=none] (179) at (3, -1.75) {};
		\node [style=none] (180) at (2.5, -1.75) {};
		\node [style=none] (181) at (3.5, -1.75) {};
		\node [style=none] (182) at (2.5, -2.75) {};
		\node [style=none] (183) at (3.5, -2.75) {};
		\node [style=none] (184) at (3, -2.25) {\tiny{$H$}};
		\node [style=none] (185) at (3, -2.75) {};
		\node [style=none] (186) at (3, -3.25) {};
	\end{pgfonlayer}
	\begin{pgfonlayer}{edgelayer}
		\draw (80.center) to (81.center);
		\draw (81.center) to (82.center);
		\draw (80.center) to (82.center);
		\draw (84.center) to (85.center);
		\draw (84.center) to (86.center);
		\draw (86.center) to (87.center);
		\draw (85.center) to (87.center);
		\draw (83.center) to (89.center);
		\draw (98.center) to (101.center);
		\draw (103.center) to (104.center);
		\draw (104.center) to (105.center);
		\draw (103.center) to (105.center);
		\draw (107.center) to (108.center);
		\draw (107.center) to (109.center);
		\draw (109.center) to (110.center);
		\draw (108.center) to (110.center);
		\draw (106.center) to (112.center);
		\draw (127.center) to (128.center);
		\draw (128.center) to (129.center);
		\draw (127.center) to (129.center);
		\draw (131.center) to (132.center);
		\draw (131.center) to (133.center);
		\draw (133.center) to (134.center);
		\draw (132.center) to (134.center);
		\draw (130.center) to (136.center);
		\draw (160.center) to (161.center);
		\draw (160.center) to (162.center);
		\draw (162.center) to (163.center);
		\draw (161.center) to (163.center);
		\draw (166.center) to (167.center);
		\draw (170.center) to (171.center);
		\draw (174.center) to (175.center);
		\draw (176.center) to (177.center);
		\draw (178.center) to (179.center);
		\draw (180.center) to (181.center);
		\draw (180.center) to (182.center);
		\draw (182.center) to (183.center);
		\draw (181.center) to (183.center);
		\draw (185.center) to (186.center);
	\end{pgfonlayer}
\end{tikzpicture}}

		\caption{}
		\label{fig:GHZState}
	\end{subfigure}
	\begin{subfigure}{0.5\textwidth}
		\centering
		
\begin{tikzpicture}
	\begin{pgfonlayer}{nodelayer}
		\node [style=none] (105) at (-0.75, -12.25) {};
		\node [style=none] (106) at (-0.75, -15.75) {};
		\node [style={white_dot, scale=1.6}] (107) at (-0.75, -11.75) {};
		\node [style=none] (110) at (-0.75, -6.25) {};
		\node [style=none] (111) at (-0.75, -11.25) {};
		\node [style=none] (112) at (2.75, -6.25) {};
		\node [style=none] (113) at (-0.25, -11.5) {};
		\node [style=none] (114) at (-2.5, -6.25) {};
		\node [style=none] (115) at (-1, -11.25) {};
		\node [style=none] (116) at (-4.25, -6.25) {};
		\node [style=none] (117) at (-1.25, -11.5) {};
		\node [style={black_dot, scale=0.3}] (118) at (0.25, -8) {};
		\node [style={black_dot, scale=0.3}] (119) at (0.75, -8) {};
		\node [style={black_dot, scale=0.3}] (120) at (1.25, -8) {};
	\end{pgfonlayer}
	\begin{pgfonlayer}{edgelayer}
		\draw (105.center) to (106.center);
		\draw [in=90, out=-90, looseness=0.75] (110.center) to (111.center);
		\draw [in=30, out=-90, looseness=0.75] (112.center) to (113.center);
		\draw [in=120, out=-90, looseness=0.75] (114.center) to (115.center);
		\draw [in=150, out=-90, looseness=0.50] (116.center) to (117.center);
	\end{pgfonlayer}
\end{tikzpicture}}

		$\mapsto$ \hspace{0.2cm}
		
\begin{tikzpicture}
	\begin{pgfonlayer}{nodelayer}
		\node [style=none] (105) at (5, -2.75) {};
		\node [style=none] (106) at (5, -12.5) {};
		\node [style={white_dot, scale=1.41}] (107) at (2.25, -9.5) {};
		\node [style={black_dot, scale=1.0}] (120) at (5, -9.5) {};
		\node [style={black_dot, scale=0.3}] (121) at (2.25, -8.25) {};
		\node [style=none] (121) at (-5.25, -7) {};
		\node [style=none] (122) at (-4.25, -8.25) {};
		\node [style=none] (123) at (-3.25, -7) {};
		\node [style=none] (125) at (-4.25, -7.5) {\tiny{$0$}};
		\node [style=none] (126) at (-2.5, -8.5) {};
		\node [style=none] (127) at (-1.5, -9.75) {};
		\node [style=none] (128) at (-0.5, -8.5) {};
		\node [style=none] (130) at (-1.5, -9) {\tiny{$0$}};
		\node [style=none] (132) at (-8, -5.25) {};
		\node [style=none] (133) at (-7, -6.5) {};
		\node [style=none] (134) at (-6, -5.25) {};
		\node [style=none] (136) at (-7, -5.75) {\tiny{$0$}};
		\node [style=none] (137) at (1.25, -10.75) {};
		\node [style=none] (138) at (2.25, -12) {};
		\node [style=none] (139) at (3.25, -10.75) {};
		\node [style=none] (141) at (2.25, -11.25) {\tiny{$0$}};
		\node [style=none] (142) at (2.25, -8.75) {};
		\node [style=none] (143) at (2.25, -10.75) {};
		\node [style=none] (144) at (1.75, -9.5) {};
		\node [style=none] (145) at (5, -9.5) {};
		\node [style=none] (146) at (-7, -2.75) {};
		\node [style=none] (147) at (-7, -5.25) {};
		\node [style=none] (148) at (-4.25, -2.75) {};
		\node [style=none] (149) at (-4.25, -7) {};
		\node [style=none] (150) at (-1.5, -2.75) {};
		\node [style=none] (151) at (-1.5, -8.5) {};
		\node [style={white_dot, scale=1.41}] (152) at (-7, -4) {};
		\node [style={black_dot, scale=1.0}] (153) at (-4.25, -4) {};
		\node [style=none] (154) at (-7.5, -4) {};
		\node [style=none] (155) at (-4.25, -4) {};
		\node [style={white_dot, scale=1.41}] (156) at (-4.25, -5.75) {};
		\node [style={black_dot, scale=1.0}] (157) at (-1.5, -5.75) {};
		\node [style=none] (158) at (-4.75, -5.75) {};
		\node [style=none] (159) at (-1.5, -5.75) {};
		\node [style={white_dot, scale=1.41}] (160) at (-1.5, -7.25) {};
		\node [style=none] (161) at (-2, -7.25) {};
		\node [style=none] (162) at (-0.5, -7.25) {};
		\node [style={black_dot, scale=0.3}] (163) at (2.25, -8) {};
		\node [style={black_dot, scale=0.3}] (164) at (2.25, -7.75) {};
		\node [style={black_dot, scale=0.3}] (167) at (0, -7.25) {};
		\node [style={black_dot, scale=0.3}] (168) at (0.25, -7.25) {};
		\node [style={black_dot, scale=0.3}] (169) at (0.5, -7.25) {};
	\end{pgfonlayer}
	\begin{pgfonlayer}{edgelayer}
		\draw (105.center) to (106.center);
		\draw (121.center) to (122.center);
		\draw (122.center) to (123.center);
		\draw (121.center) to (123.center);
		\draw (126.center) to (127.center);
		\draw (127.center) to (128.center);
		\draw (126.center) to (128.center);
		\draw (132.center) to (133.center);
		\draw (133.center) to (134.center);
		\draw (132.center) to (134.center);
		\draw (137.center) to (138.center);
		\draw (138.center) to (139.center);
		\draw (137.center) to (139.center);
		\draw (142.center) to (143.center);
		\draw (144.center) to (145.center);
		\draw (146.center) to (147.center);
		\draw (148.center) to (149.center);
		\draw (150.center) to (151.center);
		\draw (154.center) to (155.center);
		\draw (158.center) to (159.center);
		\draw (161.center) to (162.center);
	\end{pgfonlayer}
\end{tikzpicture}}

		\caption{}
		\label{fig:SpiderImplementation}
	\end{subfigure}
	\caption{\ed{The diagram in (a) shows the Euler parametrisation of a general single qubit state;} (b) shows the quantum state (a GHZ state) assigned to `that' as part of our \ansaetze\ in case of the \discocat\ model and given that $q_n=1$ and $q_s=0$; (b) depicts the quantum circuit implementation of the `merge-dot' as part of the \bow\ model given that all words are assigned single qubit states.}
\end{figure} 

For the \RP\ task, the relative pronoun `that' also appears in the vocabulary. For the baseline models this relative pronoun is treated like any other word, for the \discocat~model it however receives a special treatment. 
Note that although at the pregroup level the type of `that' depends on whether it is the subject or object case, at the quantum level, recalling that $q_s=0$ for this task, only one kind of quantum state is required. 
Following the use of \rl{the merge-dot} 
in~\shortcite{SadrzadehEtAl_2013_FrobeniusAnatomyI,SadrzadehEtAl_2014_FrobeniusAnatomyII}\footnote{\rl{As mentioned in Sec.~\ref{sec:basline_models} the merge-dot corresponds to the presence of a Frobenius algebra.}} to model functional words like relative pronouns, we chose for `that' a GHZ-state, which is displayed in Figure~\ref{fig:GHZState} and which, notably, does not involve any parameters. 

Finally, the use of \rl{the merge-dot also appeared in Section~\ref{sec:basline_models} in the definition of our \bow\ model.}
Figure~\ref{fig:SpiderImplementation} shows the implementation of that dot as part of a quantum circuit -- again, an unparametrised structure.  

With the laid out approach, the choices that fix an ansatz (with $q_n=1$ fixed) can be summarised by a triple of hyperparameters $(q_s, p_n, d)$. 
The total number $k$ of parameters, denoted  $\Theta = (\theta_1, ..., \theta_k)$, varies correspondingly with the model and depends on the vocabulary. See Tables~\ref{tab:Tab_ansaetze}  and \ref{tab:Tab_ansaetze_bl} for the \ansaetze\ we studied. 
Note that Fig.~\ref{fig:ExampleSent_QCircuit} shows the quantum circuit for the example sentence from the \MC\ task precisely for ansatz $(1,1,1)$ in the \discocat\ model.

\begin{table}[H]
\small
\centering
\begin{tabular}{|c|c||c|c|}
\hline
\multicolumn{2}{|c||}{ \MC\ } & \multicolumn{2}{|c|}{ \RP\ }\\
\hline
\hline
$(q_s, p_n, d)$ & $k_D$ & $(q_s, p_n, d)$ & $k_D$ \\[0.05cm]
\hline 
$(1,1,1)$ & 22 & $(0,1,1)$ & 114 \\[0.1cm]
$(1,1,2)$ & 35 & $(0,1,2)$ & 168 \\[0.1cm]
$(1,3,1)$ & 40 & $(0,3,1)$ & 234 \\[0.1cm]
$(1,3,2)$ & 53 & $(0,3,2)$ & 288 \\
\hline
\end{tabular}

\normalsize
\caption{Overview of the \ansaetze\ studied for the \discocat\ model, where $k_D$ is the number of parameters of the resultant model.}
\label{tab:Tab_ansaetze}
\end{table}

\color{black}

\begin{table}[H]
\small
\centering
\begin{tabular}{|c|c|c||c|c|c|}
\hline
\multicolumn{3}{|c||}{ \MC\ } & \multicolumn{3}{|c|}{ \RP\ }\\
\hline
\hline
$(q_s, p_n, d)$ & $k_W$ & $k_B$ & $(q_s, p_n, d)$ & $k_W$ & $k_B$ \\[0.05cm]
\hline 
$(1,2,1)$ & - & 34  &  $(0,1,1)$ & 116 &  - \\[0.1cm]
$(1,3,1)$ &  -  &  51 & $(0,1,2)$ & 231 & -  \\[0.1cm]
$(1,3,2)$ & 37 & -  &  $(0,2,2)$ & - & 230    \\
\hline
\end{tabular}
\normalsize
\caption{Overview of the \ansaetze\ studied for the \ws\ and \bow\ models, where $k_W$, $k_B$ are the number of parameters for the respective resultant models. Note that for the baseline models $q_s$ is irrelevant, since the only wire type that appears is $n$ (see Secs. 5.1 and 5.2), however it is still included to have a consistent notation.}
\label{tab:Tab_ansaetze_bl}
\end{table}

\subsection{Model Prediction and Optimisation}
\label{subsec:Optimisation}

Let $P$ denote a sentence in case of the \MC\ dataset or a noun phrase in case of the \RP\ dataset. 
After \textit{Step 4} of the pipeline, every such $P$ is represented by a quantum circuit according to the chosen model and ansatz. 
Let the corresponding output quantum state,\footnote{Generally, a sub-normalised state (in physics jargon).} parametrised by the set of parameters $\Theta$, be denoted $\ket{P(\Theta)}$. 
Given the models and hyperparameters studied in this work (see Tables~\ref{tab:Tab_ansaetze} and \ref{tab:Tab_ansaetze_bl}) $\ket{P(\Theta)}$ always is a single qubit state, i.e. a vector in a two-dimensional complex Hilbert space. We define 

\begin{equation}
l_{\Theta}^i(P) \ := \ \big| \hspace*{0.2cm} | \hspace*{-0.07cm} \braket{i | P(\Theta)} \hspace*{-0.05cm} |^2 - \epsilon \hspace*{0.2cm} \big| 
\end{equation}	

\noindent
where $i\in \{0,1\}$ and $\epsilon$ is a small positive number, in our case set to $\epsilon = 10^{-9}$, 
which ensures that $0 < l_{\Theta}^i(P) < 1$ and $l_{\Theta}^0(P) + l_{\Theta}^1(P) < 1$, 
so that
  
\begin{equation}
l_{\Theta}(P) :=  \frac{1}{l_{\Theta}^0(P) + l_{\Theta}^1(P)} \ \Big(l_{\Theta}^0(P), \ l_{\Theta}^1(P) \Big)
\end{equation}

\noindent
defines a probability distribution. 
The label for $P$ as predicted by the model is then obtained from rounding, i.e. defined to be $L_{\Theta}(P) := \lceil l_{\Theta}(P) \rceil$ 
with $[0,1]$ ($[1,0]$) corresponding to `food' (`IT') for the \MC\ task and to the `subject case' (`object case') for the \RP\ task.

The \MC\ dataset was partitioned randomly into subsets $\mathcal{T}$ (training), $\mathcal{D}$ (development) and $\mathcal{P}$ (testing) with cardinalities $|\mathcal{T}| = 70$, $|\mathcal{D}| = 30$, $|\mathcal{P}| = 30$. 
Similarly, for the \RP\ task with $|\mathcal{T}| = 74$, $|\mathcal{P}| = 31$, but no development set $\mathcal{D}$,  since the ratio of the sizes of vocabulary and dataset did not allow for yet fewer training data, while the overall dataset of 105 phrases could not be easily changed\footnote{In contrast to the \MC\ task, here the data was extracted from an existing dataset, and picking further phrases while ensuring a minimum frequency of all words was non-trivial (see Sec.~\ref{sec:TheTasks}).}. 
\ed{For the \RP\ task the ratio between subject to object cases was 46/28 in the training subset and 19/12 in the test subset, which reflects the ratio between the two classes in the given overall dataset. 
For the  \MC\ task the ratio between `food' and `IT' related sentences was 39/31 in the training subset, 11/19 in the development subset and 15/15 in the test subset.} 
	
The objective function used for the training is standard cross-entropy; that is, if letting $L(P)$ denote the actual label according to the data, the cost is 

\begin{equation}
C(\Theta) := - \frac{1}{|\mathcal{T}|} \sum_{P\in \mathcal{T}}  L(P)^T \cdot \text{log} \big(l_{\Theta}(P) \big)
\end{equation}
 
For the minimisation of $C(\Theta)$, the \textsc{SPSA} algorithm \cite{Spall_1998_ImplementationGuideSPSA} is used, which for an approximation of the gradient uses two evaluations of the cost function (in a random direction in parameter space and its opposite). The reason for this choice is that in a variational quantum circuit context like here, proper back-propagation requires some form of `circuit differentiation' that would in turn have to be evaluated on a quantum computer -- something being actively developed but still unfeasible from a practical perspective. The SPSA approach provides a less effective but acceptable choice for the purposes of these experiments. Finally, no regularisation was used in any form.  
\color{black}

\subsection{Classical Simulation}
\label{subsec:Simulation}

Owing to the fact that computation with NISQ devices is slow, noisy and limited at the time of writing, it is not practical to do extensive training and comparative analyses on them. This was instead done by using classical calculations to replace \textit{Steps 5-6} of Fig.~\ref{fig:pipeline} in the following sense.
For any choice of parameters $\Theta$ and some sentence or phrase $P$, the complex vector $\ket{P(\Theta)}$ can be calculated by simple linear algebra -- basically through tensor contraction. Hence the values $l_{\Theta}(P)$, and thereby also the cost $C(\Theta)$ as well as the respective types of errors, can be obtained through a `classical simulation' of the pipeline. 

\begin{figure*}[h!]
	\centering			
	\begin{subfigure}{0.47\textwidth}
		\centering
		\hspace*{-0.6cm}
		\includegraphics[scale=0.42]{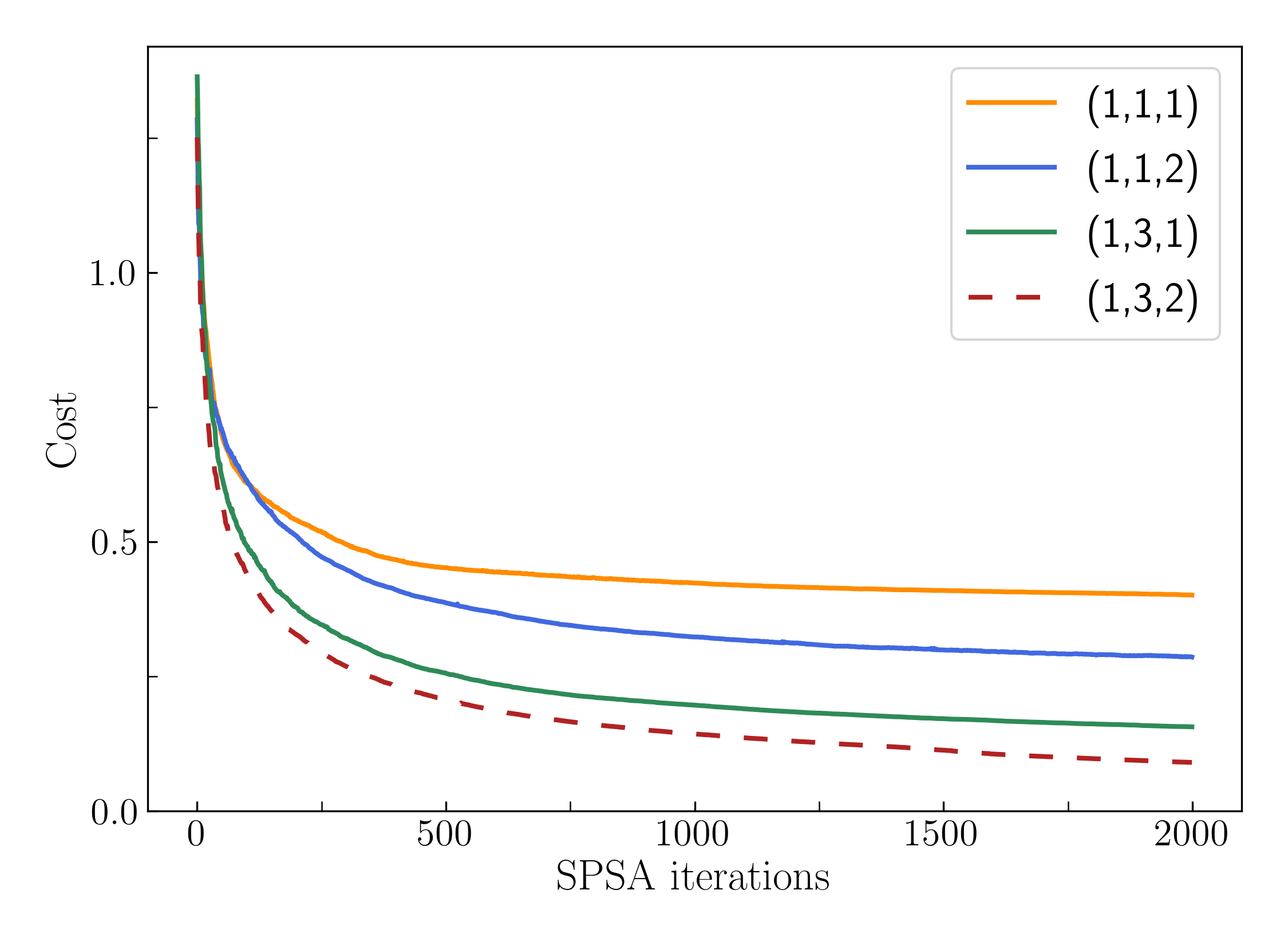}
		\vspace*{-0.6cm}
		\caption{\label{fig:MC_ClSim_overview}}
	\end{subfigure}
	\hspace*{0.5cm}
	\begin{subfigure}{0.47\textwidth}
		\centering
		\hspace*{-0.6cm}
		\includegraphics[scale=0.42]{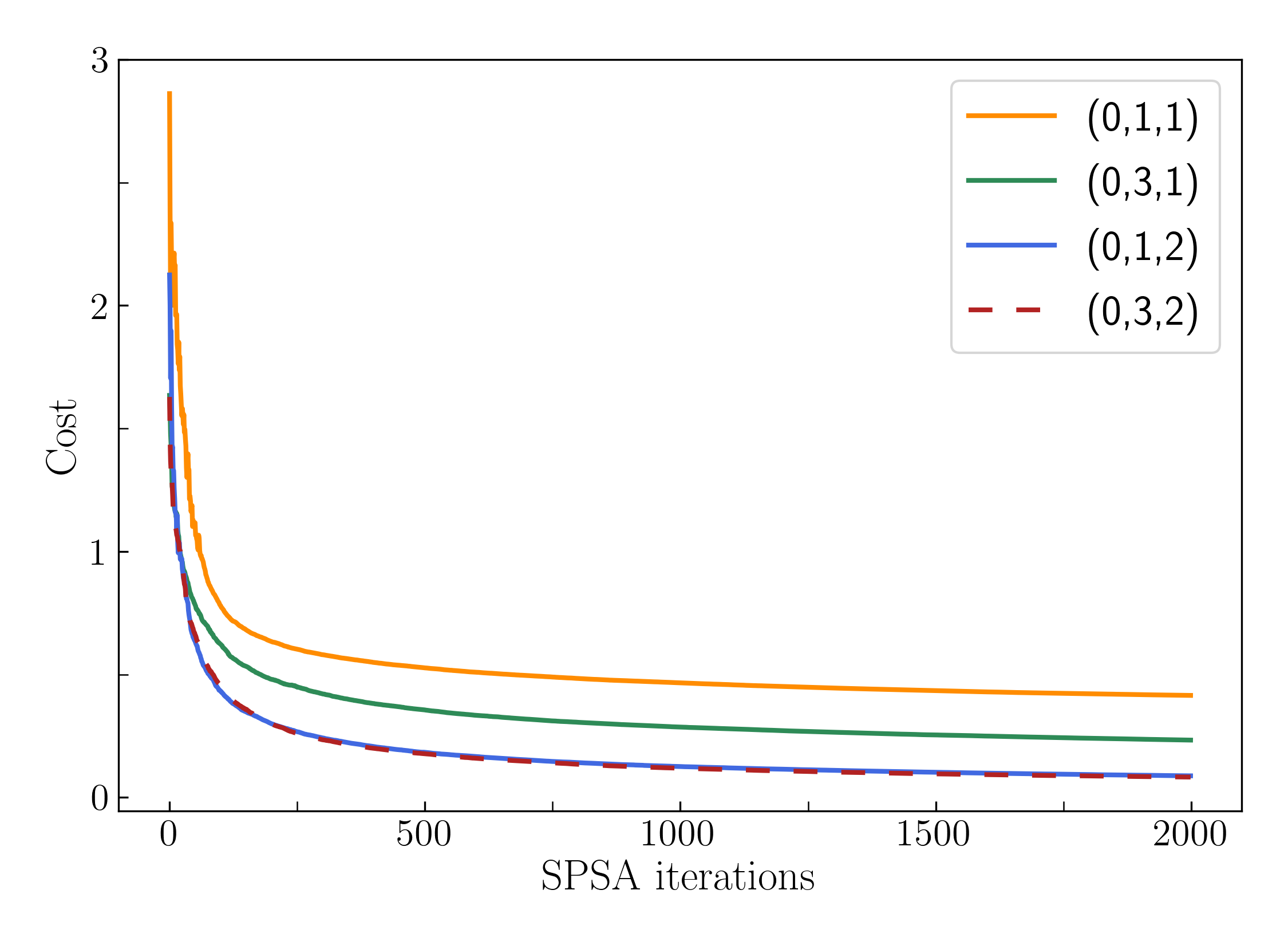}
		\vspace*{-0.6cm}
		\caption{\label{fig:RP_ClSim_overview}}
	\end{subfigure}
	\vspace*{-0.2cm}
	\caption{Convergence of the \discocat\ models in the classical simulation (averaged over 20 runs) for different \ansaetze; in (a) for the \MC\ task and in (b) for the \RP\ task.}
\end{figure*}

We therefore use this way of classically simulating the pipeline in particular to compare the different kinds of models that were introduced in preceding sections, as well as to compare the different \ansaetze\ for a fixed kind of model. Given that \discocat\ is the most complex of our models and also the main model to be studied in the actual quantum implementation, we start by comparing the different \discocat\ models that result from the different \ansaetze\ with corresponding hyperparameters as listed in Tab.~\ref{tab:Tab_ansaetze}.

Figs.~\ref{fig:MC_ClSim_overview} and \ref{fig:RP_ClSim_overview} present the convergence on the training datasets, of the \discocat\ models for the \MC\ and \RP\ task, respectively, for the selected sets of \ansaetze. 
Shown is the cost over \textsc{SPSA} iterations, where each line is the result of averaging over 20 runs of the optimisation with a random initial parameter point $\Theta$. 
The reason for this averaging is that there are considerable variances and fluctuations between any individual run due to the crude approximation of the gradient used in the stochastic \textsc{SPSA} algorithm and the specificities of the cost-parameter landscape. 
As is clear from the plots, the training converges well in all cases. 
What is more, the dependence of the minima that the average cost converges to, on the chosen ansatz reflects the theoretical understanding as follows.  

For the \MC\ task, the minimum is the lower the more parameters the model has. 
For the \RP\ task -- being about syntactic structure, essentially about word order -- the situation is different but in a way that again is understandable. 
Given our treatment of `that' (cf. Sec.~\ref{subsec:Optimisation}), it is not hard to see that the task comes down to learning embeddings such that the verbs' states become sensitive to which of their two wires connects to the first and the second noun in the phrase.   
Hence, the larger $d$ (which fixes the number of parameters for verbs) the lower the minimum. 

The plots in Figs.~\ref{fig:MC_ClSim_overview} and \ref{fig:RP_ClSim_overview} showcase what is expected from a quantum device if it were noise-free, and if many iterations and runs were feasible time-wise. 
On that basis, we chose one \discocat\ ansatz per task, for implementation on quantum hardware, that does well with as few parameters as possible:
\MCModel\ for the \MC\ task and \RPModel\ for the \RP\ task. 

It is these two selected models which we therefore focus on for the comparison with the simpler baseline models. For the \MC\ task we choose to compare the (1,3,1) \discocat\ model (40 parameters) with the (1,2,1) and (1,3,1) \bow\ models (with 34 and 51 parameters, respectively) and with the (1,3,2) \ws\ model (37 parameters). The reason for these choices is that we wish to keep a similar number of parameters compared to the \discocat\ model for a fair comparison. 

For similar reasons we choose to compare in the \RP\ task the chosen (0,1,2) \discocat\ model (168 parameters) with the (0,2,2) \bow\ model (230 parameters) and the (0,1,1) and (0,1,2)  \ws\ models (with 116 and 231 parameters, respectively.)  

\begin{figure*}[h!]
	\centering			
	\begin{subfigure}{0.47\textwidth}
		\centering
		\hspace*{-0.6cm}
		\includegraphics[scale=0.42]{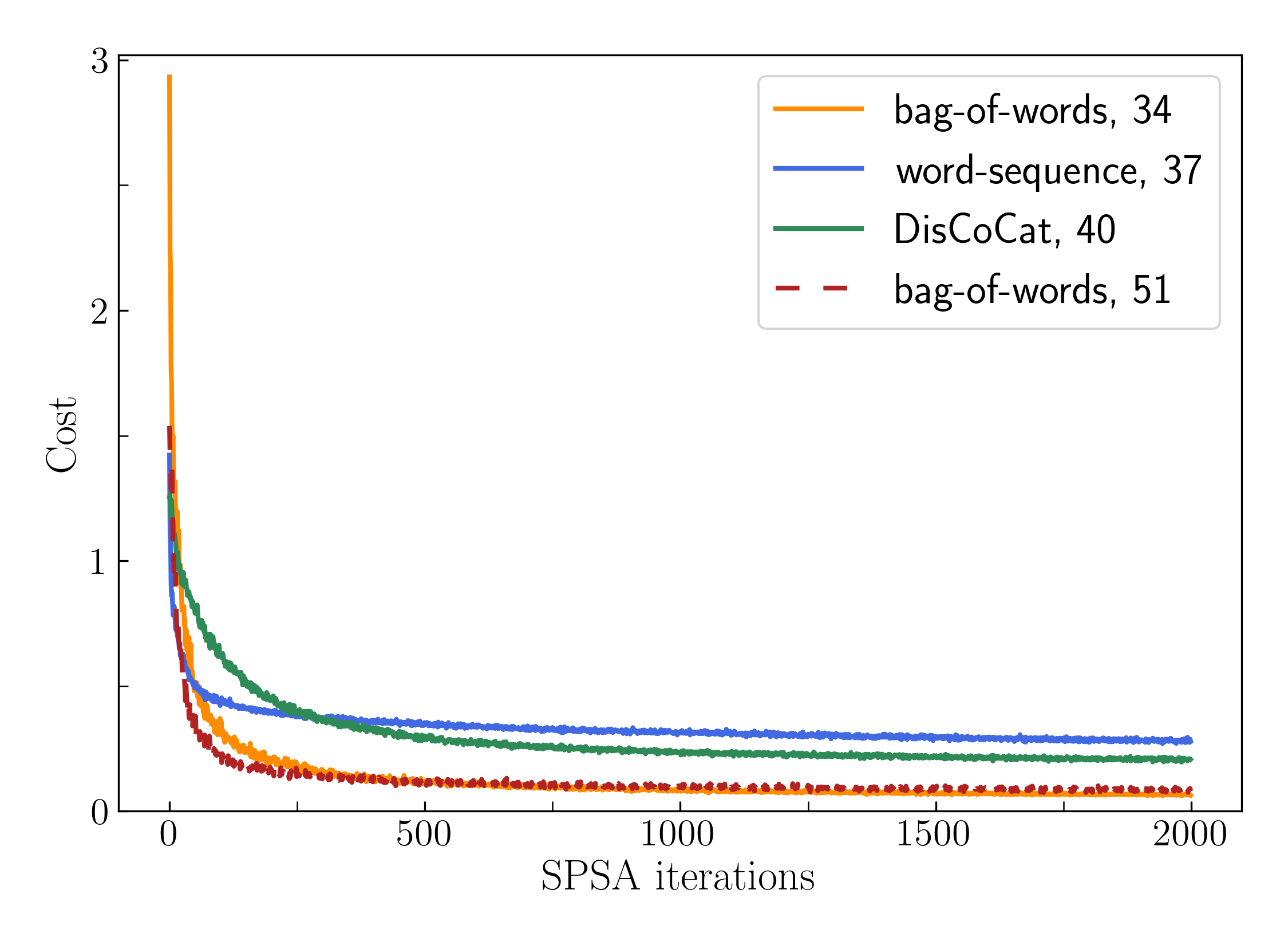}
		\vspace*{-0.6cm}
		\caption{\label{fig:MC_bl_cost}}
	\end{subfigure}
	\hspace*{0.5cm}
	\begin{subfigure}{0.47\textwidth}
		\centering
		\hspace*{-0.6cm}
		\includegraphics[scale=0.42]{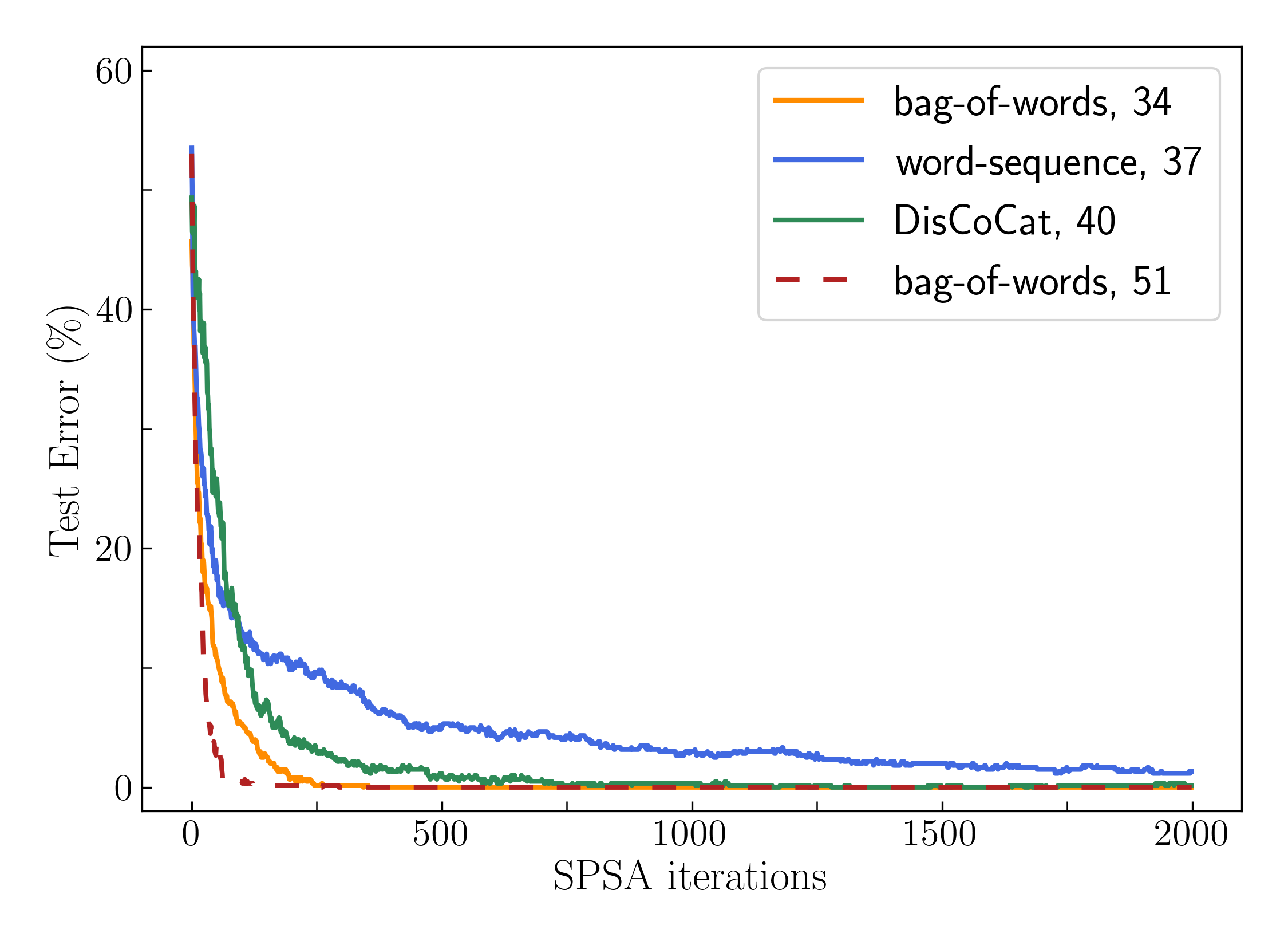}
		\vspace*{-0.6cm}
		\caption{\label{fig:MC_bl_testerr}}
	\end{subfigure}
	\vspace*{-0.2cm}
	\caption{Convergence of various models in the classical simulation (averaged over 20 runs) showing (a) the cost function and (b) the test error for the \MC\ task. \label{fig:MC_bl}}
\end{figure*}

\begin{figure*}[h!]
	\centering			
	\begin{subfigure}{0.47\textwidth}
		\centering
		\hspace*{-0.6cm}
		\includegraphics[scale=0.42]{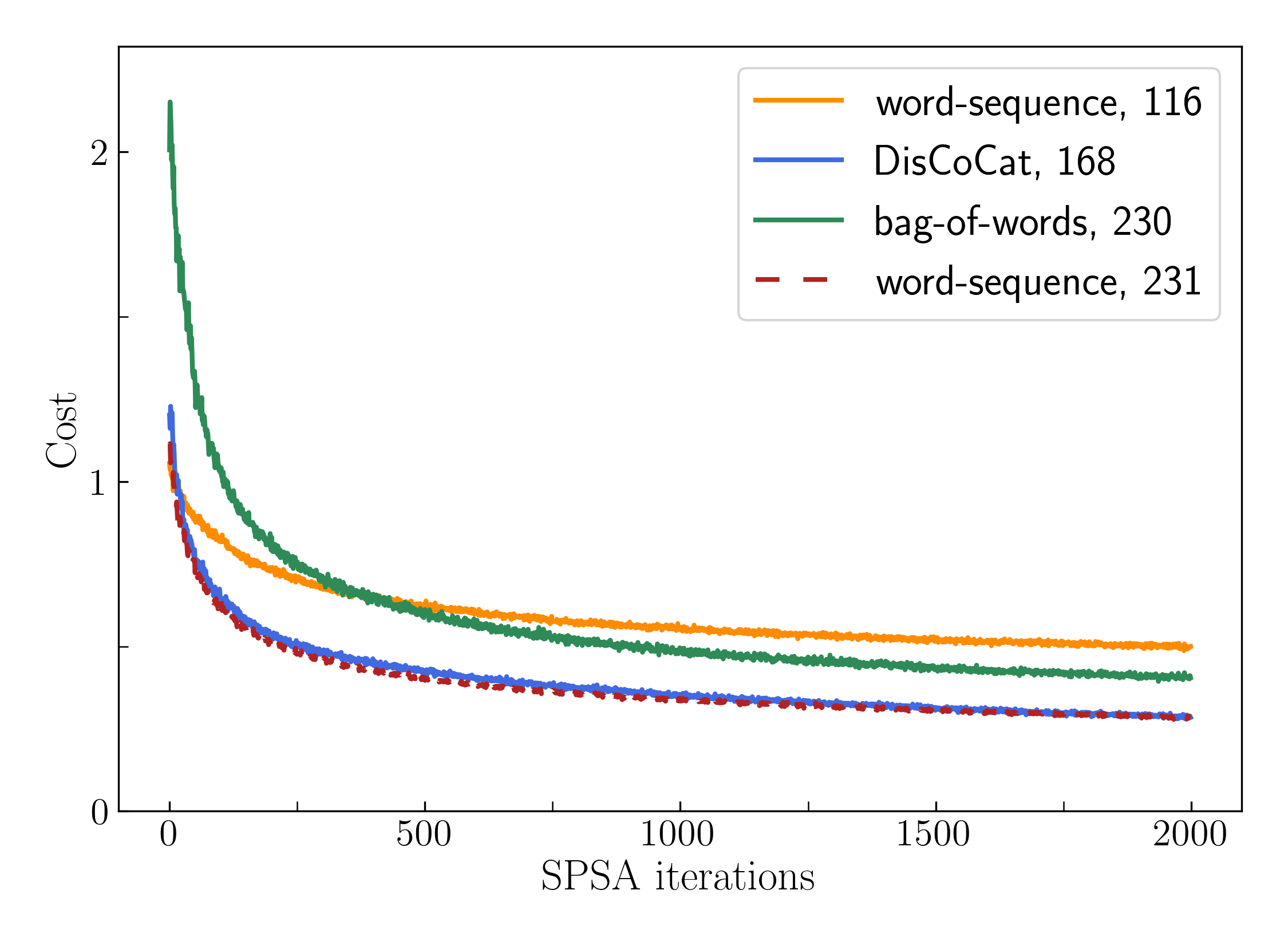}
		\vspace*{-0.6cm}
		\caption{\label{fig:RP_bl_cost}}
	\end{subfigure}
	\hspace*{0.5cm}
	\begin{subfigure}{0.47\textwidth}
		\centering
		\hspace*{-0.6cm}
		\includegraphics[scale=0.42]{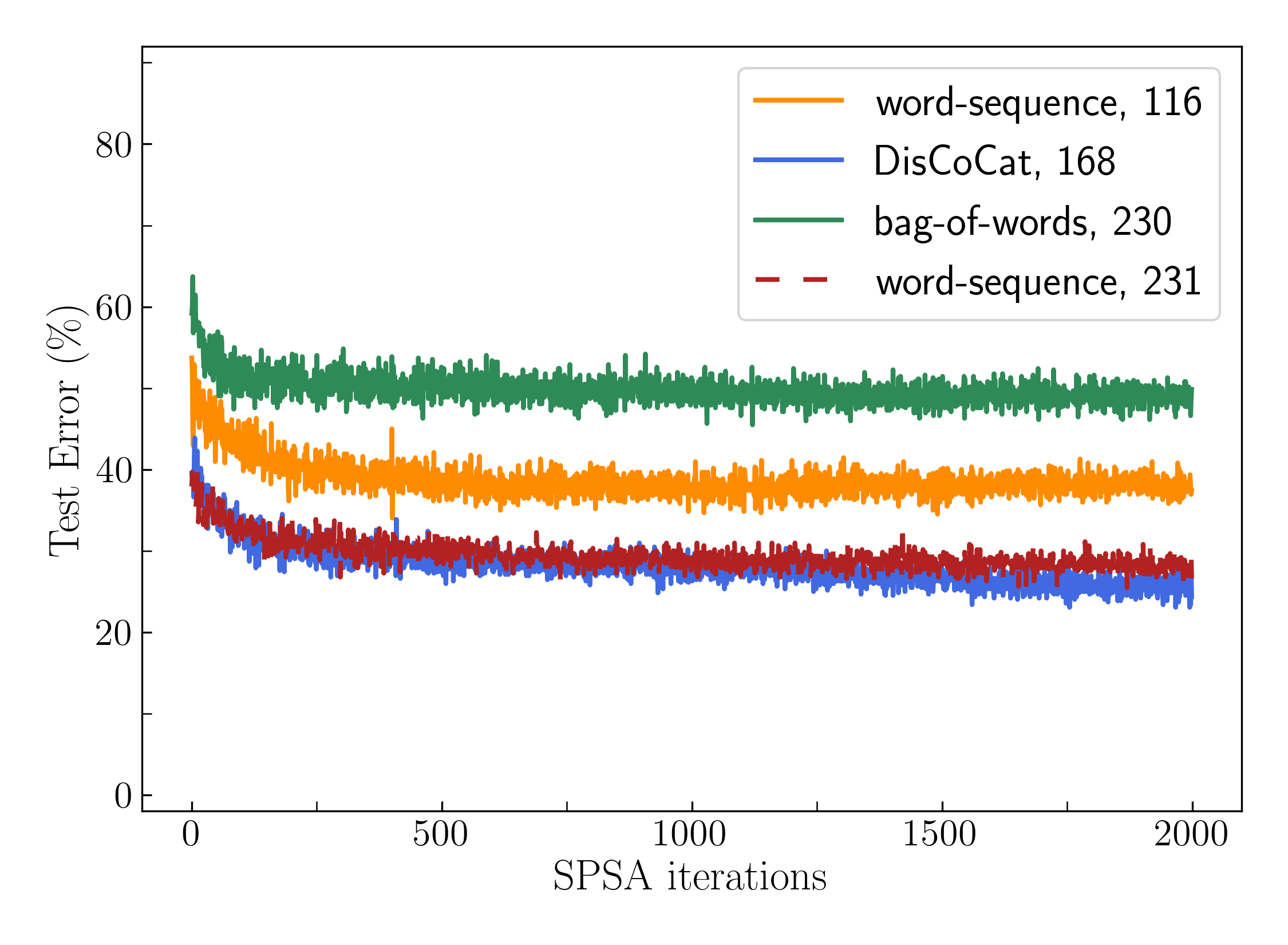}
		\vspace*{-0.6cm}
		\caption{\label{fig:RP_bl_testerr}}
	\end{subfigure}
	\vspace*{-0.2cm}
	\caption{Convergence of various models in the classical simulation (averaged over 20 runs) showing (a) the cost function and (b) the test error for the \RP\ task. \label{fig:RP_bl}}
\end{figure*}

Figure~\ref{fig:MC_bl} shows the results for the \MC\ task, and Figure~\ref{fig:RP_bl} for the \RP\ task for the comparisons outlined above. As can be seen, in the \MC\ task, being about the presence of certain words to identify the class the sentence is in, the \bow\ model does best both with more and with fewer parameters than the \discocat\ model. The \ws\ model does slightly worse than \discocat\ both in the cost function convergence for the training data and for the test errors.  

In the \RP\ task, however, being more about syntactic structure, the \discocat\ model is on a par with the 231 parameter \ws\ model (and even has a slightly lower test error than the latter), despite having only 168 parameters. Just as expected, the \bow\ model does not do better than random guessing for the test set evaluations. Also note that the \ws\ model with fewer parameters performs worse than \discocat.

\subsection{A Sanity Check}
\label{sec:sanity_check}

In light of how similarly the \ws\ and the \discocat\ model perform on the \RP\ task with its dataset suffering from biases due to its small size (relative to the size of the vocabulary), and in order to demonstrate the difference in syntax-sensitivity that the respective models should have by definition, this section presents an additional task on an entirely artificial dataset as a sanity-check of our understanding. 
The basic idea is to consider the same task as in \RP, but on a dataset that is designed to prevent the model from potentially just picking up the occurrence of certain words that signify the class, instead of actually learning the sentence order. 

The dataset was constructed on the basis of a vocabulary of 13 words with 8 nouns, 4 transitive verbs and the relative pronoun `that', to yield a perfectly balanced dataset in the following way. 
For any possible choice of a triple $(n1, n2, v)$ of two distinct nouns and one verb, all four combinatorially possible noun phrases are created that are of one of the two syntactic structures that appear in the \RP\ dataset. 
For instance, given the words `organisation' ($n1$), `teacher' ($n2$) and `support' ($v$) the four phrases are (writing $rp$ for `that'): 
\begin{center}
`organisation that support teacher' ($n1\ rp\ v\ n2$),\\[0.05cm]
`teacher that support organisation' ($n2\ rp\ v\ n1$),\\[0.05cm]
`organisation that teacher support' ($n1\ rp\ n2\ v$),\\[0.05cm]
`teacher that organisation support' ($n2\ rp\ n1\ v$).
\end{center}

Hence, each noun and each verb appear by construction an equal amount of times in the subject and the object subclause cases (in an overall dataset of size 448 noun phrases). This does of course lead to some sentences which do not make sense such as `building which like document', but the purpose of this task is not to be a linguistically interesting one --  it is only to test the performance of the models once statistical effects can be excluded. 

\begin{figure*}[h!]
	\centering			
	\begin{subfigure}{0.47\textwidth}
		\centering
		\hspace*{-0.6cm}
		\includegraphics[scale=0.42]{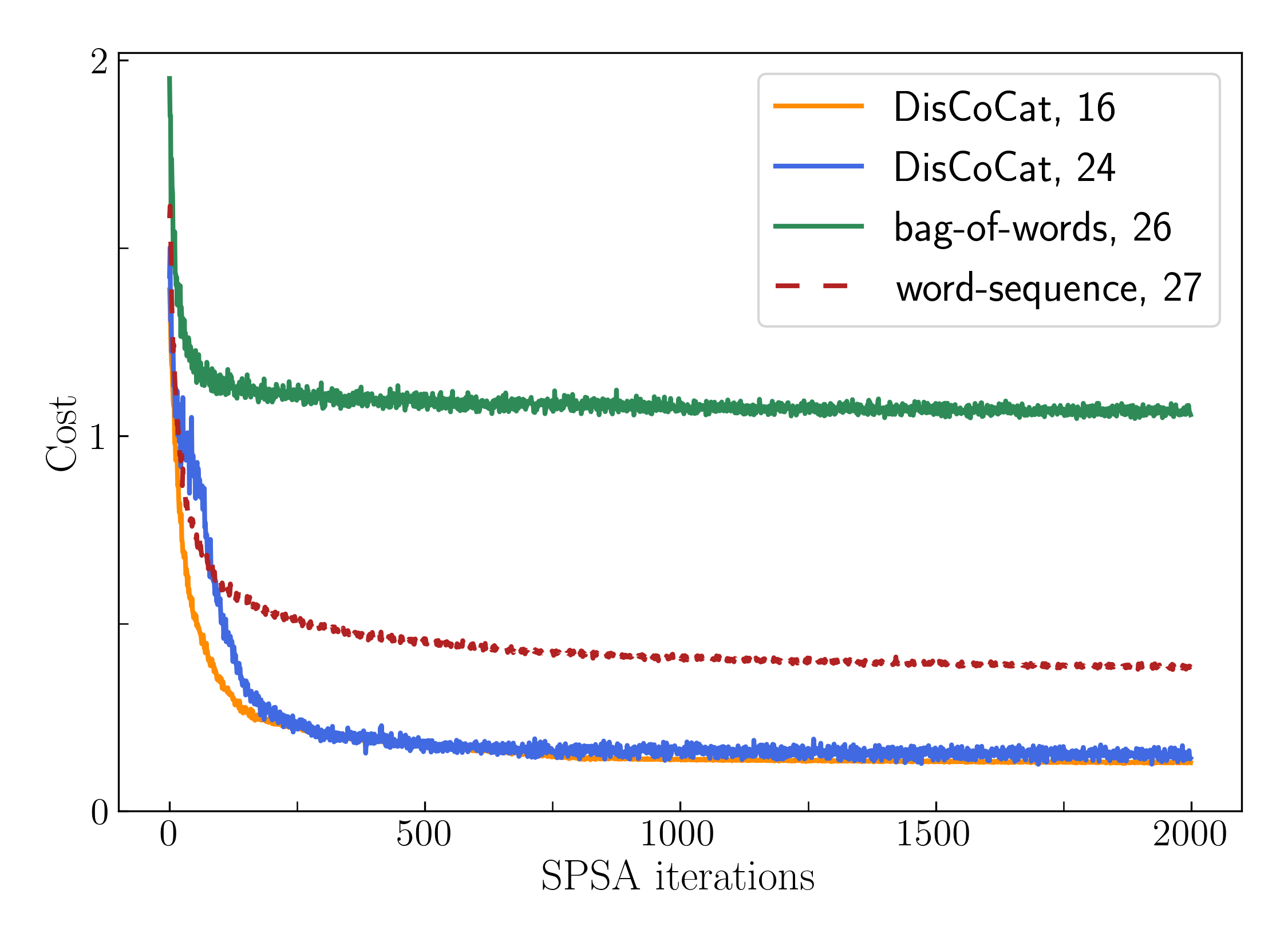}
		\vspace*{-0.6cm}
		\caption{\label{fig:RP_gen_bl_cost}}
	\end{subfigure}
	\hspace*{0.5cm}
	\begin{subfigure}{0.47\textwidth}
		\centering
		\hspace*{-0.6cm}
		\includegraphics[scale=0.42]{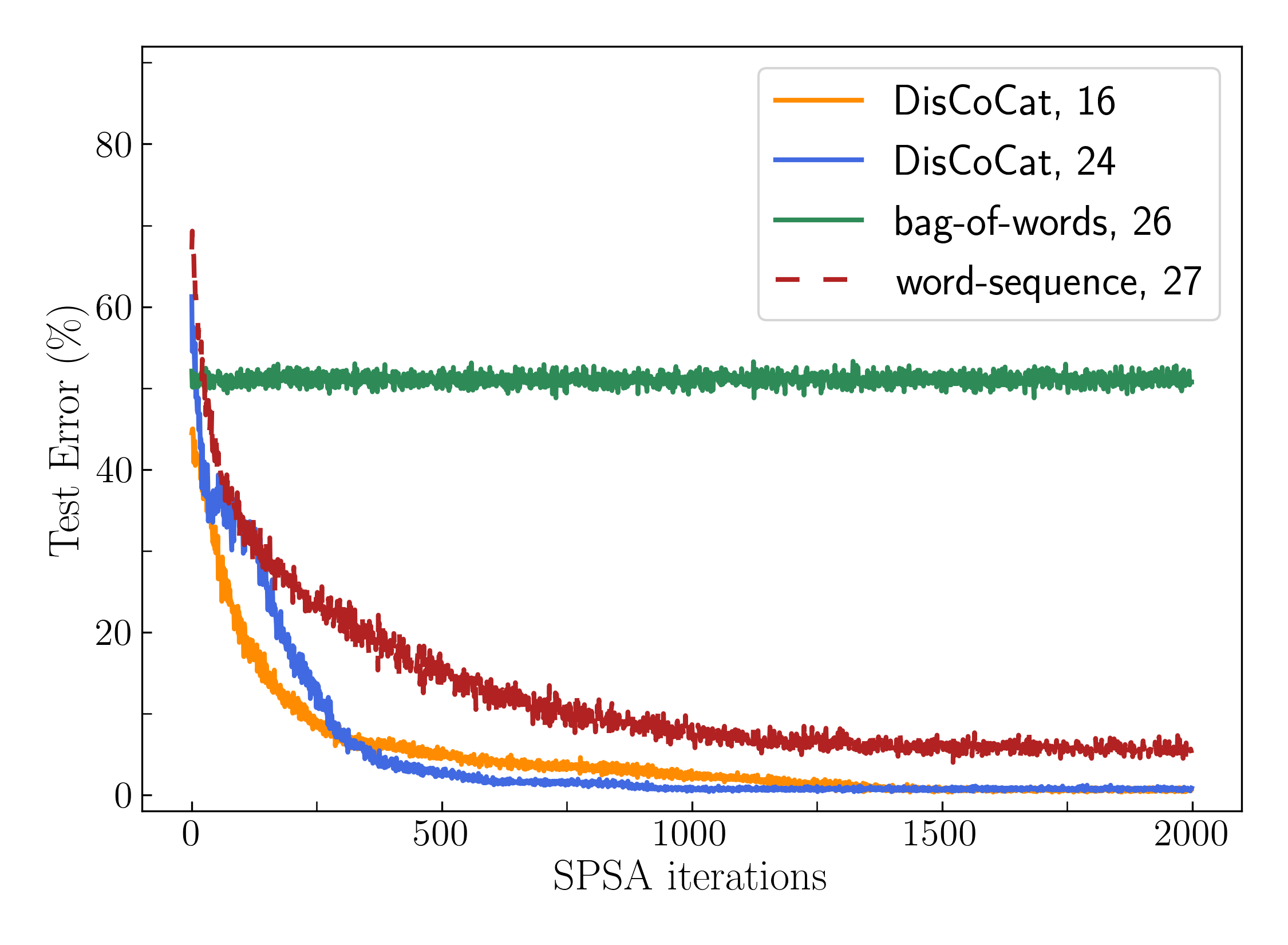}
		\vspace*{-0.6cm}
		\caption{\label{fig:RP_gen_bl_testerr}}
	\end{subfigure}
	\vspace*{-0.2cm}
	\caption{Convergence of various models in the classical simulation (averaged over 20 runs) showing (a) the cost function and (b) the test error for our sanity-check task with a large and artificial, but perfectly symmetric \RP\ like dataset. \label{RP_gen_bl}}
\end{figure*}

The models we choose to compare are the (0,1,2) \discocat\ model (with 16 parameters), 
the (0,1,4) and the (0,2,2) \bow\ models (with 24 and 26 parameters, respectively) and 
the (0,1,2) \ws\ model (27 parameters). Note that the much smaller parameter numbers are due to the very limited vocabulary in this task, compared to the original \RP\ task. The results are shown in Figure~\ref{RP_gen_bl}.

Just as expected, the \bow\ model fails completely. 
The \ws\ model does converge and learn, but is well out-performed by the \discocat\ model for both chosen \ansaetze, even though both choices of settings lead to a model with fewer parameters than the \ws\ model. 

The performances of our three models on all three tasks, \MC, \RP\ and this sanity-check task here, taken together, confirm the choice of the \discocat\ model as an economical and versatile choice of model for the experiments on actual quantum hardware.

\subsection{Quantum Runs}
\label{subsec:QuantumRuns}

\begin{figure*}
	\centering
	\begin{subfigure}{0.47\textwidth}
		\centering
		\hspace*{-0.62cm}
		\includegraphics[scale=0.44]{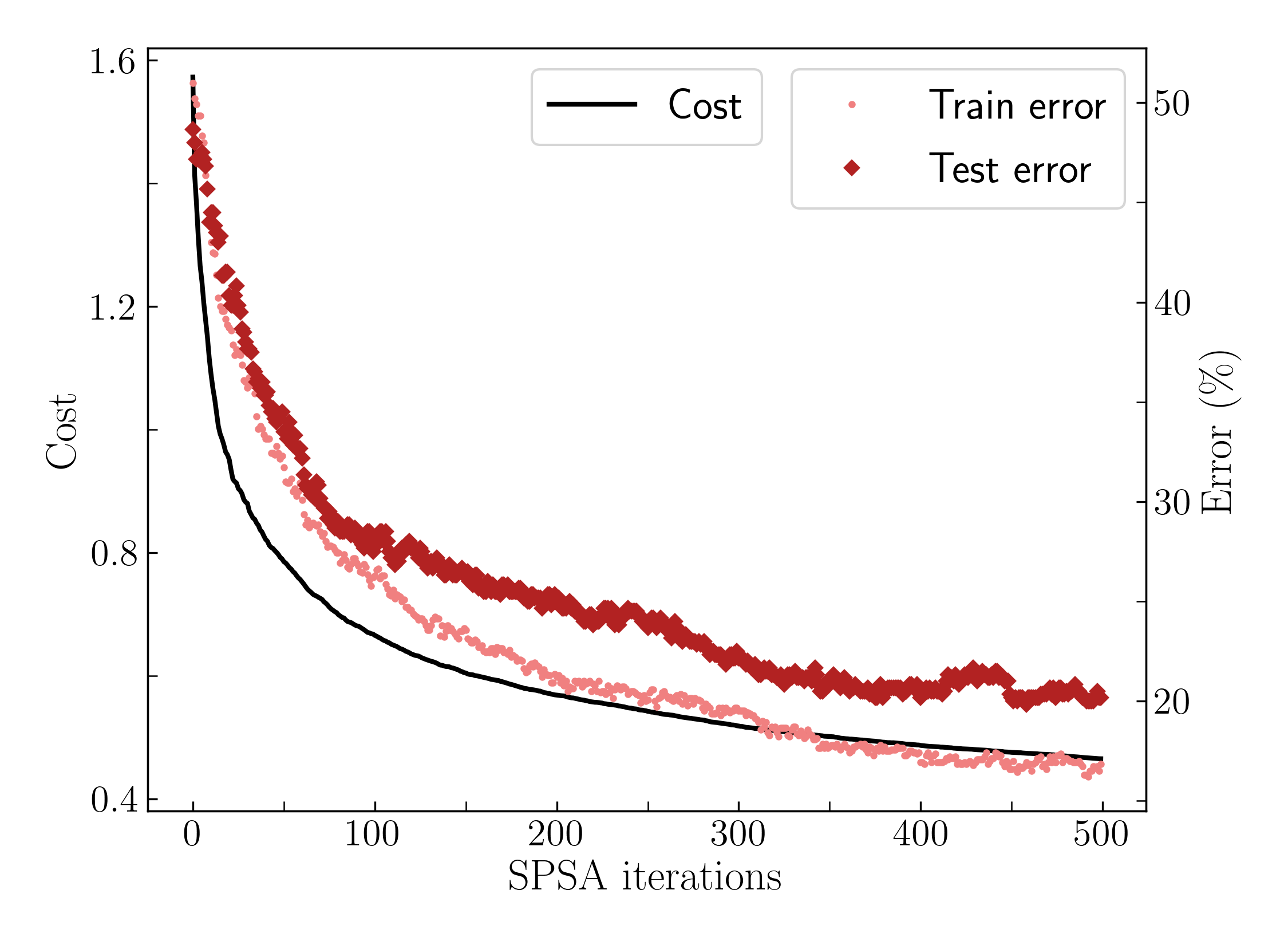}
		\vspace*{-0.6cm}
		\caption{\label{fig:MC_ClSim_results}}
	\end{subfigure}
	\hspace*{0.6cm}
	\begin{subfigure}{0.47\textwidth}
		\centering
		\hspace*{-0.6cm}
		\includegraphics[scale=0.44]{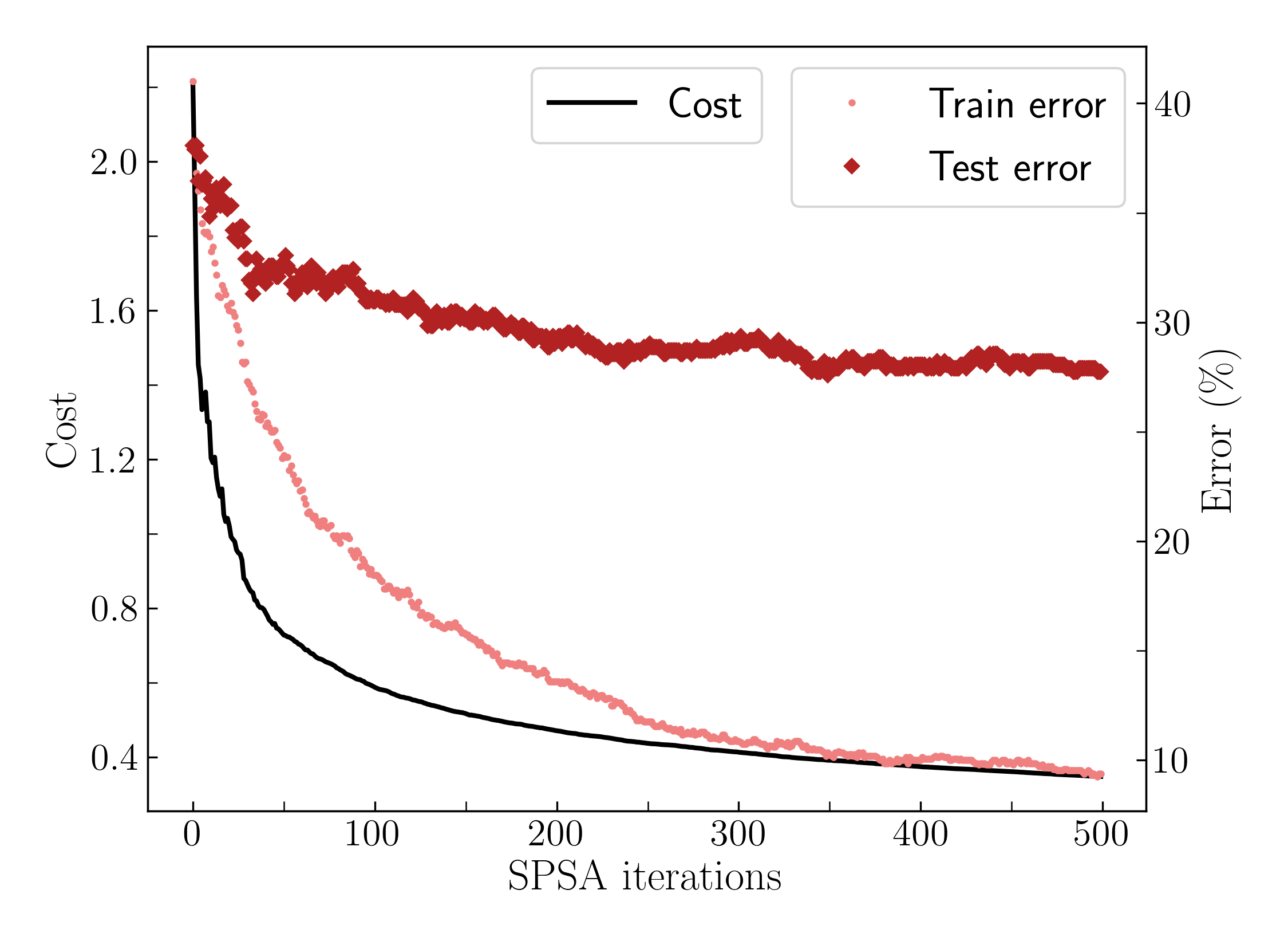}
		\vspace*{-0.6cm}
		\caption{\label{fig:RP_ClSim_results}}
	\end{subfigure}
	\vspace*{-0.2cm}
	\caption{Classical simulation results for the cost and errors (again averaged over 20 runs) for the \discocat\ model, in (a) for the \MC\ task and the chosen ansatz \MCModel\  and in (b) for the \RP\ task and the chosen ansatz \RPModel.}
\end{figure*}


\rl{We now turn to the experiments on actual quantum hardware, which concern the \discocat\ model for the two chosen \ansaetze\ for the two respective tasks (see Sec.~\ref{subsec:Simulation}). 
Note that the reason for not running the two baseline models on quantum hardware is that (especially at the time when the experiments were run) access to quantum hardware is costly and limited, in conjunction with the fact that additionally running the baseline models on quantum hardware would not yield any further scientific insight. 
The question of which models are fit for the chosen tasks can be answered through classical simulations -- see above analyses for the conclusion in favour of the \discocat\ model. 
As for studying how the pipeline performs when actually running on quantum hardware, that is, going from theory to experiment, which often incurs technical challenges and surprises, e.g., due to the noisiness, implementing the \discocat\ model suffices. 
Ignoring the conceptual differences between the latter and the baseline models, which is completely understood from simulations, the actual quantum circuits -- their structure and typical gates -- do not differ in any significant way as far as the implementability is concerned due to consistently using the same ansätze throughout.} 

\rl{Just before finally presenting the results of the quantum runs, Figs.~\ref{fig:MC_ClSim_results} and \ref{fig:RP_ClSim_results} show simulation results for the correspondingly chosen \discocat\ \ansaetze\ together with the errors, but for fewer iterations than in Figs.~\ref{fig:MC_ClSim_overview} and \ref{fig:RP_ClSim_overview} for better visibility and a more detailed comparison with the quantum runs. 
After 500 iterations} in the \MC\ case, the train and test errors read 16.9\% and 20.2\%, respectively; in the \RP\ case the train and test errors are 9.4\% and 27.7\%, respectively. 
Noticeably, the test error for the latter task is somewhat higher than the test
error for the former task. This is as expected with one of the most important reasons being the large vocabulary in combination with the small size of the dataset; for example, analysing the data in the aftermath revealed that many of the 115 words in the vocabulary appear only in the test set $\mathcal{P}$, but not at all in the training set $\mathcal{T}$.\footnote{More precisely, 17\% (36\%) of the vocabulary appear zero times (once) in $\mathcal{T}$.}

For both tasks executed on quantum hardware, all circuits (compiled with TKET$^{\text{TM}}$) were run on IBM's machine \texttt{ibmq\_bogota}.  
This is a superconducting quantum computing device with 5 qubits and quantum volume 32.\footnote{Quantum volume is a metric which allows quantum computers of different architectures to be compared in terms of overall performance, and it quantifies the largest random circuit of equal width and depth that a quantum computer can successfully implement.} 
As explained before, due to the limited access on quantum hardware, we only performed experiments with the \discocat\ model.

Every time the value of the cost or the errors are to be calculated, the compiled circuits corresponding to all sentences or phrases in the corresponding dataset ($\mathcal{T}$, $\mathcal{D}$ or $\mathcal{P}$) are sent as a single job to IBM's device. 
There, each circuit is run $2^{13}$ times (the maximum number permitted at the time).
The returned data thus comprises for each circuit (involving $q$ qubits) $2^{13} \times q$ measurement outcomes (0 or 1).  
\ed{As explained in detail in Sec.~\ref{sec:pipeline} (also see Sec.~\ref{sec:Intro_QC} for the general idea) the pipeline then involves for every sentence or phrase $P$ appropriately post-selecting the data by restricting the measurement data to the 0 outcomes on all those qubits that feature a 0-effect in $P$'s circuit diagram.\footnote{Note that the \MC\ dataset includes sentences that lead to circuits on only three qubits. Here `appropriately post-selecting' means post-selecting two out of the three used qubits.}   
Once post-selected, the relative frequencies of outcomes 0 and 1 of the remaining qubit that carries the output state of $P$ (corresponding to the wire without a 0-effect in $P$'s circuit diagram), give the estimate of $|\braket{i | P(\Theta)} |^2$ (with $i=0,1$) and thus of $l_{\Theta}(P)$.}
The remaining post-processing to calculate the cost or an error is then as for the classical simulation. 

\begin{figure*}
	\centering
	\begin{subfigure}{0.47\textwidth}
		\centering
		\hspace*{-0.62cm}
		\includegraphics[scale=0.44]{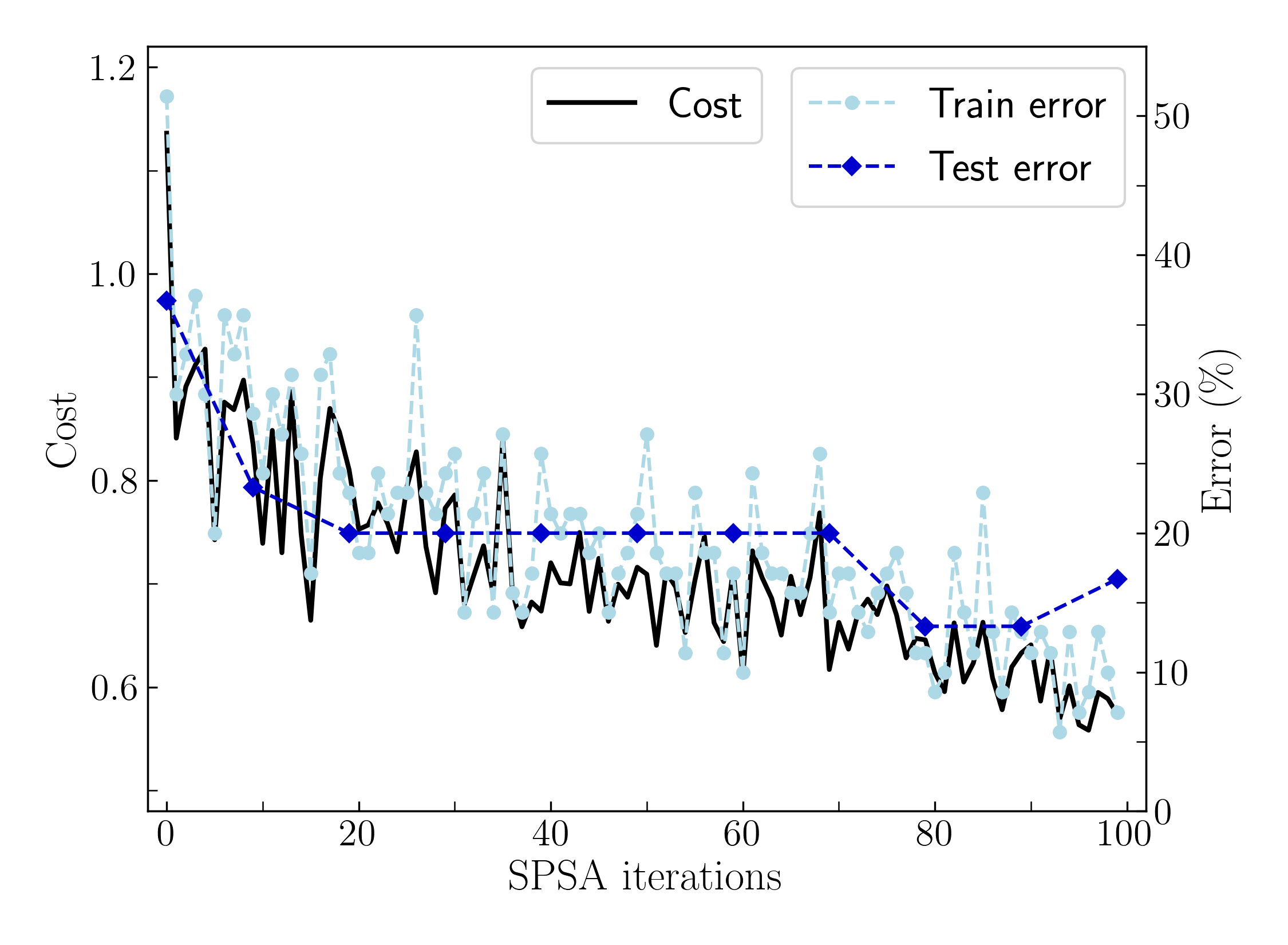}
		\vspace*{-0.6cm}
		\caption{\label{fig:MC_Quantum_Results}}
	\end{subfigure}
	\hspace*{0.6cm}
	\begin{subfigure}{0.47\textwidth}
		\centering
		\hspace*{-0.58cm}
		\includegraphics[scale=0.44]{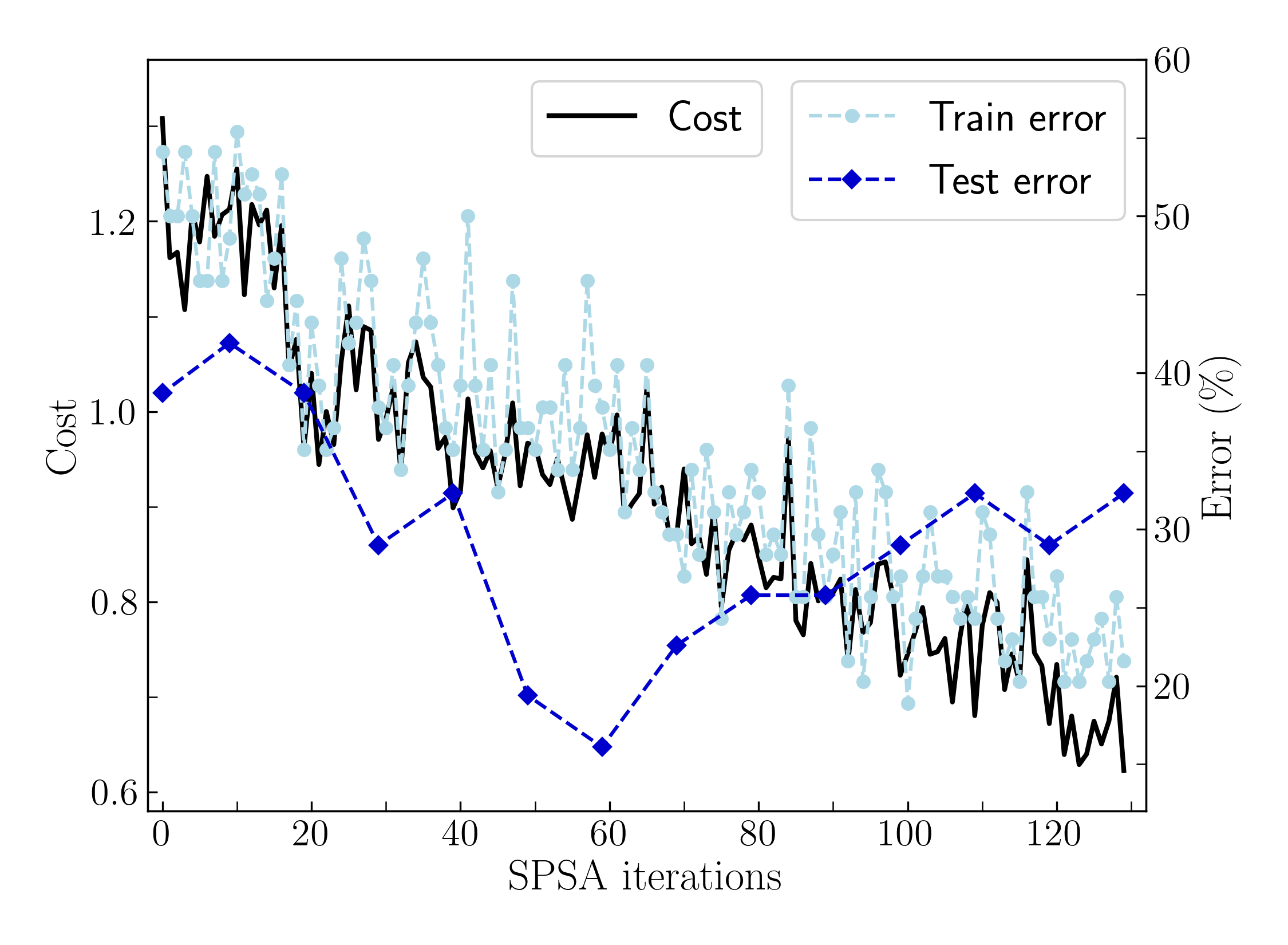}
		\vspace*{-0.6cm}
		\caption{\label{fig:RP_Quantum_Results}}
	\end{subfigure}
	\vspace*{-0.2cm}
	\caption{Results from quantum computation for the cost, as well as train and test errors (test error for every 10th iteration) for the \discocat\ model, in (a) for the \MC\ task and chosen ansatz \MCModel\ and in (b) for the \RP\ task and chosen ansatz \RPModel.}
	\label{fig:Quantum_Results}
\end{figure*}

The experiments involved one single run of minimising the cost over 100 iterations for the \MC\ task and 130 iterations for the \RP\ task, in each case with an initial parameter point that was chosen on the basis of simulated runs on the train (and dev) datasets.\footnote{This choice was made to reduce the chances of being particularly unlucky with the one run that we did on actual quantum hardware. Yet, this choice's significance should not be overrated given that the influence of quantum noise spoils the predictability of the cost at a particular parameter point from simulated data.} 
For the \MC\ task, obtaining all the results shown in Fig.~\ref{fig:MC_Quantum_Results} took just under 12 hours of run time. This was enabled by having exclusive access to \texttt{ibmq\_bogota} for this period of time. 
In contrast, the \RP\ jobs were run in IBMQ's `fairshare' mode, i.e. with a queuing system in place that ensures fair access to the machine for different researchers. 
As a consequence, for the \RP\ task, which is not computationally more involved than the \MC\ task, obtaining all the results shown in Fig.~\ref{fig:RP_Quantum_Results} took around 72 hours. With access to quantum devices still being a limited resource and `exclusive access' being rationed, we see here the reason for the problems that the time cost of yet larger datasets would entail.

Figures~\ref{fig:MC_Quantum_Results} and \ref{fig:RP_Quantum_Results} show the cost and various errors for the \MC\ task with ansatz \MCModel\ and for the \RP\ task with ansatz \RPModel, respectively. 
Despite the noise levels that come with NISQ era quantum computers, and given the fact of a single run with few iterations compared to the classical simulations in Sec.~\ref{subsec:Simulation}, the results look remarkably good -- the cost is decreasing with SPSA iterations in both cases, modulo the expected fluctuations. 

After 100 (130) iterations as reported in Fig.~\ref{fig:MC_Quantum_Results} (Fig.~\ref{fig:RP_Quantum_Results}), the test error was 16.7\% (32.3\%) for the~\MC~and~\RP ~task, respectively, with F-score 0.85 (0.75). These results were checked to be statistically significant against random guessing with $p\leq 0.001$ for~\MC~and $p \leq 0.10$ for~\RP~according to a permutation test.

Compared to the simulations (Figs.~\ref{fig:MC_ClSim_results}, \ref{fig:RP_ClSim_results}), it can be seen that after the same number of iterations, test errors are actually lower for the quantum runs. However, due to the special conditions under which these experiments were performed (single run on quantum hardware subject to quantum noise versus many averaged runs on classical hardware without noise, but with the inherent instability of SPSA optimization still present), such comparisons are not very conclusive. In general, the trends presented in the plots of Figure \ref{fig:Quantum_Results} are the expected based on the size of the datasets. For example, the test error in Fig.~\ref{fig:RP_Quantum_Results} shows a paradigmatic example of overfitting around iteration 60.

\section{Future Work and Conclusions}
\label{sec:Conclusion}

In this work we have provided a detailed exposition of two experiments of NLP tasks implemented on small noisy quantum computers. Our main goal was to present novel larger-scale experiments building on prior proof-of-concept work \cite{meichanetzidis2020grammaraware}, while having in mind the AI/NLP practitioner. We provided for the first time quantum versions of three compositional models, each entailing a different degree of syntax sensitivity, and we tested them successfully on well-defined, although simple in nature, NLP tasks.
Despite the prototypical nature of the currently available, albeit rapidly growing in size and quality, quantum processors, we obtain meaningful results on our datasets. 
We also hope that the current exposition will serve as a useful introduction to practical QNLP for the AI/NLP community.

Having established a QNLP framework for near-term quantum hardware,
we briefly outline directions for future work.
The ansatz circuits we have used to parameterise the word meanings served well for this work's goal and also are motivated in the QML literature by the fact that they are conjectured to be hard to simulate classically. 
However, it was beyond the scope of this work to search for optimal word-circuits in a task-specific way. 
This opens up an exploratory arena for future work on \ansaetze.
In particular, an open question regards trade-offs of performance of ansatz families in a specific task versus general performance on many tasks.

\ed{Furthermore, a crucial direction for further work regards scalability and resource estimation.
There is of course more than one way that one can think of scaling up QNLP tasks.
What is special to QNLP is how scaling up in different dimensions manifests itself as different resource costs in the context of quantum computation, especially given the modest quantum devices available today.
First of all, we can consider the cost as the sentences get longer.
As a sentence scales in length, the number of qubits on which its corresponding quantum circuit is defined, i.e. the circuit width, will scale as well, depending on the number of qubits assigned to each pregroup type.
This consideration is remedied by the realisation that quantum computers have been growing in qubit numbers and there is no sign of this growth slowing down.
More importantly, however, a longer sentence will incur an exponential time-cost in the number of qubits being post-selected. 
Note that in the long term, one does not aim to post-select, but employ more sophisticated protocols where only one qubit needs to be measured, resulting in additive approximations of an amplitude encoding a tensor contraction \cite{arad2008quantum}.
Of course, in natural language, sentences are usually upper bounded in length anyway so that we can consider these as up-front constant costs.} 

\ed{Moreover, the \discocat\ framework is applicable to other typological grammars.
For example, parsing of a sentence with a CCG grammar returns a syntax tree decorated with grammatical types. For such tree structures one can again define a `syntactic functor', as done in this work, and instantiate them as quantum circuits for NLP tasks. Experimenting with other grammars  than pregroup grammars opens up a playground for defining a whole landscape of QNLP models.
Our {\tt lambeq} Python package serves as the platform to begin exploring the possible species of compositional QNLP models.
Also note that in our current setup, the number of parameters that one needs to optimise scales polynomially with the size of the vocabulary.
This motivates the careful study of the landscapes defined by a task's cost function, as well as the exploration of other optimisation methods beyond SPSA.}  
\ed{Another important direction is that we aim to experiment with pre-trained quantum word embeddings,
instead of training word-states from scratch for each task, as done in this work.
These can then be used in any downstream QNLP task and their training would be considered an upfront cost. We expect that using pre-trained quantum word embeddings would be beneficial, seeing as it has been shown to be beneficial to employ pre-trained word embeddings in classical NLP pipelines.} 

\ed{Hence, all of the above ideas and developments taken together open up the possibility of readily using large-scale textual data in NLP tasks relevant to the real world, moving away from the proof-of-concept setting of this work, whose motivation was only to introduce this novel approach to both the quantum computing and NLP communities.}

\ed{Before closing, worthy of a comment is the apparent linearity of the model studied in this work. 
Indeed, quantum theory is a linear theory, at least as far as the unitary evolution of pure states is concerned.
However, the subtlety lies in how one chooses to embed the input data, in this case the word meanings,
and how these embeddings enter the cost function. 
In this work the word embeddings are defined as pure quantum states
and the labels that enter the cost function are defined in terms of the probability distribution over the measurement outcomes for the particular qubit that carries a sentence's or phrase's representation. 
Importantly, the Born rule, which gives the probabilities, is a non-linear function of the amplitudes defined in terms of the pure states.
More generally, quantum machine learning with variational circuits
can be viewed elegantly in terms of kernel methods \shortcite{SchuldEtAl_2019_QMLInFeatureHilbertSpace,schuld2021quantum}.
In this light, it becomes clear that the mapping from the parameters defining the input data to the cost is non-linear.} 

\ed{Relating this then to a potential quantum advantage, a possible avenue for obtaining a quantum advantage arises when a QNLP task is designed so that the evaluation of the cost function is hard to simulate classically due to a high degree of entanglement and interference, all the while demonstrating better performance than the classical state of the art methods.
Another avenue to quantum advantage that we aim to investigate is defining compositional models such that any variational optimisation
in the pipeline involves circuits that are small enough to be efficiently simulatable classically, but also such that when these components are composed to form a larger quantum-circuit-based model, a quantum computer would be needed to evaluate it and perform the QNLP task at hand. 
These types of quantum advantage in the field of NLP would be `meaningful' in that they would constitute examples of non-contrived real-world applications of quantum computers in the near-term.}

It is with a view to such advantages that experimenting with NLP models on quantum computing hardware matters now, so that we are prepared for the technological breakthroughs as they happen. There is always a gap between theory and practice and while there is a considerable body of work on the theoretical side of things, practical experimental implementations remain largely unexplored within QNLP. Quantum computers are here to stay and it is thus imperative to start exploring what they can offer the field of NLP and, more generally, AI.

\section*{Acknowledgments}

We are grateful to Richie Yeung for his help on technical issues, and also, along with Alexis Toumi, for \textsc{DisCoPy} support.
We also thank Marcello Benedetti for helpful discussions.
\rl{We would furthermore like to thank Quantinuum's TKET$^{\text{TM}}$ team for support with \texttt{PyTKET}.} 
We acknowledge the use of IBM Quantum services for this work. The views expressed are those of the authors, and do not reflect the official policy or position of IBM or the IBM Quantum team. 

\bibliography{ref}
\bibliographystyle{theapa}

\end{document}